\documentclass[12pt,oneside]{article}
\newcommand{\figs}{}
\newcommand{\Path }{}
\usepackage{\Path Paper_style}
\usepackage{\Path Keywords}
\usepackage{\Path Environments}
\usepackage{tikz-cd}
\usepackage{tabularx} \newcolumntype{L}{>{\raggedright\arraybackslash}X}
\usepackage{Paper_keywords}
\renewcommand{\blue}[1]{ #1 }

\begin{document}
	\title{\vspace{-2cm} Conditional expectation using compactification operators }
	\author{Suddhasattwa Das\footnotemark[1]}
	\footnotetext[1]{Department of Mathematics and Statistics, Texas Tech University, Texas, USA}
	\date{\today}
	\maketitle
	\begin{abstract} The separate tasks of denoising, least squares expectation, and manifold learning can often be posed in a common setting of finding the conditional expectations arising from a product of two random variables. This paper focuses on this more general problem and describes an operator theoretic approach to estimating the conditional expectation. Kernel integral operators are used as a compactification tool, to set up the estimation problem as a linear inverse problem in a reproducing kernel Hilbert space. This equation is shown to have solutions that allow numerical approximation, thus guaranteeing the convergence of data-driven implementations. The overall technique is easy to implement, and their successful application to some real-world problems are also shown.
		\\\emph{MSC 2020 classification} : 46E27, 46E22, 62G07, 62G05
		\\\emph{Keywords} : Markov kernel, statistical denoising, compact operators, RKHS
	\end{abstract}
	
	\section{Introduction} \label{sec:intro}
	
	In many experiments, due to the uncertainty in the parameters of the setup, the outcome has to be interpreted as a conditional expectation. This is owing to the fact that the measured outcome which is ideally the ``typical" or ``true" outcome, is in fact conditional on the prevailing parameters. This notion of conditional expectation has different interpretations in various contexts, such as mean-curves, least square fitting, and denoising. We present an operator theoretic approach to the problem of finding conditional expectation, which also provides a robust technique for denoising. The technique uses ideas from both kernel mean embedding as well as kernel-based operator compactification. It has an easy adaptation to be data-driven, which we also prove is convergent in the limit of large data.
	
	To give our discussion a more firm mathematical footing we make the following assumption : 
	
	\begin{Assumption} \label{A:1}
		There is a compact metric space $X$ and a topological space $Y$, each equipped with their Borel $\sigma$-algebras $\Sigma_X$ and $\Sigma_Y$ respectively.  There is a probability measure $\mu$ on the product space $\left( X\times Y, \Sigma_{X \times Y} \right)$. 
	\end{Assumption}
	
	We interpret $X$ as the space being directly observed, and $Y$  to be the space from which a random input or parameter is drawn. The observation or measurement being performed is via the following function :
	
	\begin{Assumption} \label{A:2}
		There is an unknown function $f \in X\times Y\to \real$ which lies in the space $C(X\times Y)$, and is integrable with respect to $\mu$.
	\end{Assumption} 
	
	Our focus is on the conditional expectation of such functions :
	\begin{equation} \label{eqn:def:Ex_f}
		\bar{f} := \Ex^{\mu}(f) \in L^1(\mu_X), \quad \bar{f}(x) := \int_{y\in Y} f(x, y) d\mu(y|x) .
	\end{equation}
	In spite of the simplicity of \eqref{eqn:def:Ex_f}, $\bar{f}$ has a problem of not being well defined at every point on $X$. In general, $\bar{f}$ is only an $L^1$ equivalence class and has no guarantee of having a continuous representative, without further assumptions. We later introduce some technical but broad assumptions which would enable these conditional expectations to be continuous functions. At the moment we look at some examples.
	
	\paragraph{Examples} The situation described in Assumptions \ref{A:1} and \ref{A:2} occurs commonly in many situations.
	\begin{enumerate}
		\item Additive noise : Consider the situation where $\bar{f}$ is a random variable (r.v.) on $X$, the space $Y$ is $\real^{p}$ endowed with a zero mean distribution $\mu_Y$, and $f(x,y) = \bar{f}(x) + y$ is a contamination of $\bar{f}$ with noise $y$ \citep[e.g.][]{dangeti2003denoising, elad2006image}. Let the product space $X\times Y$ be endowed with the product measure $\mu = \mu_X \times \mu_Y$. Then the task of denoising is about recovering $\bar{f}$,  which is related to $f$ and $\mu$ via \eqref{eqn:def:Ex_f}.
		\item Pull-backs : Suppose $\paran{ \Omega, \tilde{\mu} }$ is a probability measure, and $\calX:\Omega\to X$ and $\calY:\Omega\to Y$ are two random variables. Let $H_{\calX}$ be the subspace of $L^2(\mu)$ consisting of functions of the form $ \SetDef{ \phi\circ \calX }{ \phi\in L^2 \paran{ \calX_* \mu } } $. Thus these are the square integrable functions which factorized through $\calX$. Alternatively, these are the pullbacks of square integrable functions on $X$, under $\calX$. A space $H_{\calX \times \calY}$ can be defined similarly. 
		
		Now set $\mu = (\calX\times \calY)_* \tilde{\mu}$, the push forward of $\tilde{\mu}$ onto $X\times Y$. Then for any function $f\in L^2(\mu; \real)$, the pullback $f\circ \paran{ \calX \times \calY }$ lies in $H_{\calX \times \calY}$. Then 
		\[ \paran{\Ex^{\mu} f} \circ \calX = \mathbb{E} \paran{ f\circ (\calX\times \calY) | \calX } = \proj_{ H_{\calX} } \paran{ f\circ \paran{ \calX \times \calY } }. \]
		Thus the conditional expectation from \eqref{eqn:def:Ex_f} pulls back to an equality of random variables. This situation is the focus of the \emph{input model uncertainty} -problem in statistics. The two r.v.-s $A, B$ represent sub-systems of the larger system $\Omega$, and $f$ is a statistic depending on the outcomes of $A$ and $B$. Then $\mathbb{E} \paran{ f\circ (A\times B) | A }$ can be interpreted to be the mean value of the statistic $f$, as the input parameter $B$ is varied. The second equality in the equation above also implies that the conditional expectation may be derived as a least squares estimate, with a proper choice of norm.
		\item Manifold learning : The notion of principal curves and manifolds are used to describe formulate manifold learning within a statistical context \citep[e.g.][]{izenman2012intro, bhatt2012ext, pat2015nonp}. Principal curves capture the notion of a curve passing through the center of a distribution. While there is no unique definition, it mostly relies on an expectation minimizing function from a manifold $M$. In Section ~\ref{sec:example}, we convert this expectation minimization problem into a conditional expectation problem, by assuming an unknown prior distribution from which the data-points are generated. 
	\end{enumerate}
	
	In Section~\ref{sec:example} we investigate a few real world manifestations of these above scenarios, using the theoretical and numerical tools that we build. 
	Despite its importance in several applications, robust estimation techniques are yet to be fully explored. We next try to understand the challenges associated to this task.
	
	\paragraph{Challenges} There are multiple objectives one needs to be careful about in any such technique :
	\begin{enumerate}[(i)]
		\item Smoothness : the estimated conditional expectation function preferably has some degree of regularity.
		\item Consistency : the outcome of the estimation technique should converge to the truth with more data.
		\item Data-driven : ideally the technique should not assume any prior distribution.
		\item Robustness : The problem with trying to approximate the integral in  \eqref{eqn:def:Ex_f} is that there may not be sufficient number of samples along each leaf. The techniques should have some robustness to this undersampling problem.
	\end{enumerate}
	
	We now take a brief look at a few important paradigms developed to address this estimation problem. Most of them lack in addressing one or more out of the above four objectives. 
	
	\paragraph{Related work} While Assumptions \ref{A:1} and \ref{A:2} describe a basic scenario in many theoretical and real-world situations, they do not provide a recipe for estimating $\bar{f}$ from \eqref{eqn:def:Ex_f}. There has been several ideas proposed, which we organize into five classes. Nearest-neighbor based techniques \citep[e.g.][]{li2009nonparametric, duval2010parameter, frosio2018statistical} are data-driven and easily scalable with large data, but lack a framework to guarantee consistency. The idea is to denoise a local pirece of the data cloud by taking an average of nearest neighbors, via a technique known as Non-Local-Means (NLM) \cite{duval2010parameter, frosio2018statistical}. In spite of different weighting schemes being proposed, these techniques also suffer from the presence of bias. A second class of techniques use principal component analysis and its statistical properties \citep[e.g.][]{hastie1989principal}. However, these techniques are framed in a set of Assumptions more restrictive than our general case. A third important class of techniques are based on the notion of \emph{principal curves} \citep[e.g.][]{gerber2013regularization}. Here, the target function is set to be a curve passing through the middle of a distribution. One then uses any of the vast number of gradient-descent-based techniques to minimize the squared divergence from the mean curve.  Principal curve estimation techniques come with all the advantages and disadvantages of gradient-descent learning, and also lack a firm footing in probabilistic assumption. \blue{Also as shown in \cite{duchamp1996extremal}, principal curves often turn out to be saddle points of the mean-squared projection distance. This means they are not local minima and local minima only along some selected directions of descent.} A fourth class of techniques rely on concepts from traditional Harmonic analysis. The most notable is a technique named \emph{conditional mean embedding} \cite{song2009hilbert}, which sets up the estimation as a linear-algebraic concept. Another idea is to assume a hypothesis space, and a noise prior on the coefficients with respect to a basis or frame \citep[e.g.][]{mihcak1999low, hamza2001image}. 
	
	\blue{A subclass of this fourth class of techniques are kernel based methods. These utilize the theory of kernel integral operators and RKHS to the extent that they form a class on their own. We make special note of the idea in \cite{li2022optimal} of using kernel ridge regression to estimate the conditional mean embedding.} Some other notable examples are the use of kernels for non-isotropic mixing spatial data
	\cite{lu2002spatial}, using local linear kernel estimators \cite{hallin2004local}, or adaptation of ideas from kernel density estimation \cite{steckley2003kernel}. 
	
	Our proposed technique falls under this fifth class. We use a combination of compactification and reproducing kernel Hilbert space theory to perform our estimation.
	
	\paragraph{Outline} We describe the technical details next in Section~\ref{sec:tech}. We setup the problem in a hypothesis space called reproducing kernel Hilbert space. The target function is to be determined via a linear inverse problem. To make the problem robust to finite rank approximations, we use compact operators on both sides of the equation. Readers looking for the algorithmic implementation and convergence guarantees are directed to the next Section~\ref{sec:numeric}. We demonstrate some applications in Section~\ref{sec:example} and provide some discussions in Section~\ref{sec:conclus}. Finally, we provide the proofs of our theorems and lemmas in Section~\ref{sec:proofs}.

	\section{The technique} \label{sec:tech}
	
	We begin by being more specific about the measure $\mu$. Let $\Prob(X)$ and $\Prob(Y)$ respectively denote the space of probability measures on the measurable spaces $(X, \Sigma_X)$ and $(Y, \Sigma_Y)$. \blue{Given metric spaces $Z, Z'$, $C(Z; Z')$ denote the space of continuous maps from $Z$ to $Z'$, endowed with the compact-open topology. This space is also a metric space with respect to the supremum norm. When $Z'=\real$, we denote $C(Z;\real)$ as simply $C(Z)$. Taking $Z=X$ and $Z'=C(Y)$ gives the metric space $C \paran{X; C(Y)}$. This is the space of all $\real$-valued functions on $X\times Y$ which are continuous with respect to the $X$-variable in the $C(Y)$-norm.} 
	
	For a general measure $\mu \in \Prob(X\times Y)$, the conditional measures $\SetDef{ \mu\rvert_{x} }{ x\in X }$ are guaranteed to exist only up to a set of $\mu_X$ measure zero.  Under additional assumptions on $\mu$ and $f$ we get the following very important regularity result on conditional expectations : 
	
	\begin{lemma} \label{lem:djhe09}
		Suppose Assumptions \ref{A:1} holds, and $f$ be a function in $C(X; C(Y))$. Further suppose that there is a probability measure $\mu_X\in \Prob(X)$, and a continuous map $m : \support(\mu_X) \to \Prob(Y)$. This leads to a probability measure $\mu$ on $\paran{X\times Y, \Sigma_{X \times Y}}$, defined as 
		\[ \mu(A) := \int_{x\in X} m \SetDef{ y\in Y }{ (x,y)\in A } d\mu_X(x) , \quad \forall A \in \Sigma_{X \times Y}.  \]
		Then the conditional expectation \eqref{eqn:def:Ex_f} can be realized as a function in $C \paran{ \support(\mu_X) }$. 
	\end{lemma}
	
	Lemma~\ref{lem:djhe09} is proved in Section~\ref{sec:proof:djhe09}. We next introduce our main tool : \emph{kernel functions}.
	
	\paragraph{Kernel} A kernel on the space $X$ is a bivariate function $k:X\times X \to \real$, which is supposed to be a measure of similarity between points on $X$. Bivariate functions such as distance and inner-products are examples of kernels. Kernel based methods offer a non-parametric approach to learning, and have been used with success in many diverse fields such as spectral analysis \cite{DasGiannakis_delay_2019, DasGiannakis_RKHS_2018}, discovery of spatial patterns \citep[e.g.][]{GiannakisDas_tracers_2019, DasDimitEnik2020}, and the discovery of periodic and chaotic components of various real world systems \citep[e.g.][]{DasMustAgar2023_qpd, DasEtAl2023traffic}, and even abstract operator valued measures \cite{DGJ_compactV_2018}. We shall use the widely used \emph{Gaussian} kernels, defined as
	\begin{equation} \label{eqn:def:GaussK}
		k_{\text{Gauss}, \delta} (x,y) := \exp \paran{ -\frac{1}{\delta} \dist( x, y )^2 } , \quad \forall x,y \in X. 
	\end{equation}
	Gaussian kernels have been shown to have deep connections with the geometry or topology of the underlying space \citep[e.g.][]{CoifmanLafon2006, TrillosSlepcev2018, BDGV_spectral_2020, BerrySauer_locker_2016, VonLuxburgEtAl2008}. Gaussian kernels have the important property of being \emph{strictly positive definite}, which means that given any distinct points $x_1, \ldots, x_N$ in $X$, numbers $a_1, \ldots, a_N$ in $\real$, the sum $\sum_{i=1}^{N} \sum_{i=1}^{N} a_i a_j k(x_i, x_j)$ is non-negative, and zero iff all the $a_i$-s are zero. 
	
	Closely associated to kernels are kernel integral operators (k.i.o.). Given a probability measure $\nu$ on $X$, one has an integral operator associated to a continuous kernel $k$, defined as
	\[ K^\nu : L^2(\nu) \to C^r(X) , \quad (K^\nu \phi)(x) := \int_X k(x,y) \phi(y) d\nu(y) . \]
	If the kernel $k$ is $C^r$, then its image set will also be $C^r$ functions. For this reason, k.i.o.-s are also known as smoothing operators. In fact, under mild assumptions, k.i.o.-s embed functions in $L^2(\nu)$ into function spaces of higher regularity, called \emph{RKHS}. Recall that a kernel $k$ is \emph{symmetric} if for every $x,x'\in X$, $k(x,x') = k(x',x)$. Symmetric kernels allow the use of tools from RKHS theory, which we review shortly.
	
	\paragraph{ \blue{Localized kernels} } A parameterized family of kernels $k_\delta$  is said to be \textit{uniformly localizable} \citep[][]{BerrySauer_locker_2016, BDGV_spectral_2020} with respect to a reference measure $\beta$ if there are constants $\gamma > 2 + \dim(\beta)$ such that$\in M$
	\begin{equation} \label{eqn:def:local_ker}
		\abs{k_\delta( x,x')} = \bigO{\left( \frac{ \dist(x,x')}{\delta} \right)^{-\gamma}} \; \mbox{ as } \delta\to 0^+ , \quad \forall x,x'\in X.
	\end{equation}
	Guassian kernels such as \eqref{eqn:def:GaussK}, \eqref{eqn:def:GaussSymm} have an exponential decay and are uniformly localizable with respect to any reference measure. Let $K_\delta^\beta$ be the associated reference measure. Then 
	\[\sup_{x \in X} \abs{ \paran{K_\delta^\beta f}(x) - \int_{ \SetDef{x'}{\dist(x,x')<R} } k_{\delta}(x,x') f(x') d\beta(x')} = \norm{f}_{L^\infty} \bigO{ (\delta/R)^\beta } , \quad \forall f\in L^\infty \paran{ mu^{(X)} } . \]
	This indicates that at any point of continuity of $\bar{f}$, the Markov operator-smoothed function $\Smooth_\delta^{\mu_X} \bar{f}$ will be a close approximation of $\bar{f}$. The comparison of $\Smooth_\delta^{\mu_X} \bar{f}$ with $\bar{f}$ is a separate and vast topic of its own, and there are myriad set of conditions \citep[][e.g.]{CoifmanLafon2006, HeinEtAl2005, BerryEtAl2013, ZhaoGiannakis2016, BerrySauer2016, VaughnBerryAntil2019} under which the former approximates the latter in various choices of metric. The techniques that we develop (Theorems \ref{thm:1}, \ref{thm:2} and Algorithm \ref{algo:1}) aim to approximate $\Smooth_\delta^{\mu_X} \bar{f}$ and not the true conditional expectation  $\bar{f}$. The motivation behind this is that $\Smooth_\delta^{\mu_X} \bar{f}$ being a continuous function allows pointwise comparisons, which $\bar{f}$ does not. For choice of a smoothing kernel, any kernel satisfying \eqref{eqn:def:local_ker} for $\beta = \mu_X$ would be sufficient. If the kernel decays faster than any polynomial, it wold work for every $\mu$. 
	
	\paragraph{Convolution of kernels.} Suppose $\beta$ is a fixed measure, and $k_1, k_2$ are two kernels on $X$. Then these two kernels can be combined in a procedure called \emph{convolution} to get a kernel 
	\[ \paran{ k_1 \star_{\beta} k_2 } \in C\paran{ X\times X; \real }, \quad \paran{ k_1 \star_{\beta} k_2 }(x,z) := \int k_1(x, y) k_2(y, z) d\beta(y) .  \]
	Let $K_1^{\beta},K_2^{\beta}$ be the integral operators corresponding to these kernels. Then note that
	\[\begin{split}
		\paran{ K_1^{\beta} K_2^{\beta} \phi }(x) &= \int \int k_1(x, y) k_2(y, z) \phi(z) d\beta(z) d\beta(y) = \int \int k_1(x, y) k_2(y, z) \phi(z) d\beta(y) d\beta(z) \\
		&= \int \paran{ k_1 \star_{\beta} k_2 }(x,z) \phi(z) d\beta(z)
	\end{split}\]
	Thus the operator that we get by composing two kernel integral operators is also a kernel integral operator, whose kernel is the convolution of the kernels of the two operators being composed.

	\paragraph{RKHS} A reproducing kernel Hilbert space or \emph{RKHS} is a Hilbert space of continuous functions, in which pointwise evaluations are bounded linear functionals. Any continuous symmetric, strictly positive definite kernel $k$ (such as \eqref{eqn:def:kdiff}) induces an RKHS which contains linear sums of the form $\sum_{n=1}^N a_n k(\cdot, x_n) $, in which the inner product is given by
	\[ \bracketBig{ \sum_{n=1}^N a_n k(\cdot, x_n) , \sum_{m=1}^M b_m k(\cdot, y_m) }  = \sum_{n=1}^{N} \sum_{m=1}^{M} a_n^* b_m k(x_n, y_m).  \]
	The full details of the construction of this space $\rkhs$ can be found in any standard literature \citep[e.g.][]{Paulsen2016}. The functions $k(\cdot, x_n)$ are called the \emph{sections} of the kernel $k$. The kernel sections are members of the RKHS and span the RKHS. One of the defining properties of RKHS is the \emph{reproducing} property :
	\[ \bracketBig{ k(\cdot, x) , f } = f(x) , \quad \forall x\in X, \, \forall f\in \rkhs . \]
	When an RKHS is used as the hypothesis space in a learning problem, the target function is assumed to be a finite sum $\sum_{n=1}^N a_n k(\cdot, x_n) $ of the kernel sections. Let $\nu$ be any probability measure on $X$, and $K^\nu$ be the kernel integration operator associated to $k$ and $\nu$. Then it is well known that the image of $K^\nu$ lies in $\rkhs$. We denote this image as $\rkhs_\nu$. For example, let $\nu$ be a discrete measure $\nu = \sum_{n=1,2,\ldots} w_n \delta_{x_n}$, i.e., an aggregate of Dirac-delta measures supported on discrete points $x_n$ along with weights $w_n> 0$ which sum to $1$. Then $\rkhs_\nu$ is precisely the span of the kernel sections $\SetDef{ k(\cdot, x_n) }{ n=1,\ldots,N }$. 
	The theory of RKHS remains in the background of our work. Our key idea is more a simple application of the theory of the decomposition of a measure into its conditional measures. We describe this next.
	
	\paragraph{Kernel smoothing} Suppose $\alpha$ is a probability measure on $X$, absolutely continuous with respect to $\mu_X$. Then :
	\[\begin{split}
		\int_X \bar{f}(x) d \alpha(x) &= \int_X \int_Y f(x, y) d\mu(y|x) d\alpha(x) = \int_X \int_Y f(x, y) d\mu(y|x) \frac{d\alpha}{d\mu_X}(x) d \mu_X(x) \\
		&= \int_X \int_Y f(x, y) \frac{d\alpha}{d\mu_X}(x) d\mu(y|x) d \mu_X(x) \\
		&= \int_{X\times Y} \SqBrack{f(x, y) \frac{d\alpha}{d\mu_X}(x) } d\mu(x,y) .
	\end{split}\]
	By this trivial manipulation, an integral of $\bar{f}$ which we consider to be unknown, against the measure $\alpha$ which we also assume to be unknown, is converted into an integral over the joint domain $X\times Y$. We now repeat this idea not for a single such absolutely continuous probability measure $\alpha$, but for an entire parameterized family. Such a family of absolutely continuous probability measures is realized by a Markov integral operator, built from a \emph{Markov transition kernel} as defined below :
	\[ p : X\times X\to \real_0^+, \quad \forall x\in X, \, \int p(x,x') d\mu_X(x') = 1. \]
	Note that for every $x\in X$, the kernel section $p(x,\cdot)$ is a non-negative function with integral equal to one. Thus it can be interpreted as a probability density. 
	Then we have an associated integral operator $P^{\mu_X} : L^2(\mu_X) \to C(X)$, whose action on any $\psi\in L^2(\mu_X)$ is given by
	\[ \paran{P^{\mu_X} \psi}(z) := \int_X p(z, x) \psi(x) d\mu(x) = \bracketBig{ p_*(z) , \psi }, \quad \forall z\in X. \]
	An important kernel for us will be a Markov normalized version of the Gaussian kernel \eqref{eqn:def:GaussK}. Given a measure $\beta$ on $X$ and a bandwith parameter $\delta>0$, define
	\begin{equation} \label{eqn:def:GaussSymm}
		k_{ \text{Gauss}, \delta }^{ \text{symm}, \beta } : X \times X \to \real , \quad k_{ \text{Gauss}, \delta }^{ \text{symm}, \beta }(x, x') := k_{ \text{Gauss}, \delta } (x, x') / \int_X k_{ \text{Gauss}, \delta } (x, x'') d\beta(x'').
	\end{equation}
	We denote the associated integral operator by $\Smooth_\delta^{\beta}$. Thus
	\[ \paran{ \Smooth_\delta^{\beta} \phi }(x) := \SqBrack{ \int_X k_{ \text{Gauss}, \delta } (x, x'') d\beta(x'') }^{-1} \int k_{ \text{Gauss}, \delta } (x, x') \phi(x') d\beta(x) . \]
	Given any Markov kernel $p$, the composite operator $P^{\mu_X} \Smooth_\delta^{\mu_X}$ has as its kernel the convolved kernel 
	\[ \tilde{p}_\delta := p \star_{\mu_X} k_{\text{Gauss}, \delta} .\]
	This kernel $\tilde{p}_\delta$ on $X$ has the following trivial extension to a transition kernel
	\[ q_\delta : X \times (X\times Y) \to \real_0^+, \quad q_\delta(x,x'.y) := \tilde{p}_\delta(x,x') = \int p(x, x'') \exp \paran{ -\frac{1}{\delta} \dist^2( x'', x' ) } d\mu_X(x''). \]
	\blue{Although $q_\delta$ is declared a function of three variables, we keep it independent of the third variable $y$.} While $q_\delta$ is not a kernel in a true sense, it still generates integral operator-like action
	\[ Q_\delta^{\mu} : L^2(\mu) \to C(X), \quad \paran{Q^{\mu}_\delta \phi}(x) := \int_{X\times Y} q_\delta(x,x',y) d\mu(x',y) . \]
	This trivial extension allows us to write
	\begin{equation} \label{eqn:scheme}
		P^{\mu_X} \Smooth_\delta^{\mu_X} \bar{f} = P^{\mu_X} \Smooth_\delta^{\mu_X} \Ex^{\mu} f = Q_\delta^{\mu} f .
	\end{equation}
	The simple identity in \eqref{eqn:scheme} underlines our main idea. It is based on the compactness of integral operators.
	
	\paragraph{Approximations of measures} From a practical point of view, neither of the measures $\mu$ or $\mu_X$ are known explicitly. Instead they would be approximated by measures $\alpha$ and $\nu$ respectively, which we assume satisfies the following :
	
	\begin{Assumption} \label{A:4}
		There are two probability measures $\alpha$ and $\nu$ supported on $X\times Y$ and $X$ respectively, such that $\nu$ is absolutely continuous with respect to $\alpha_X := \paran{\proj_{X}}_* \alpha$, the projection of $\alpha$ to $X$. Moreover, $d \nu / d \alpha$ is a function bounded away from zero.
	\end{Assumption}
	
	The measure $\alpha$ is meant to be an approximator of $\mu$. Similar to \eqref{eqn:scheme}, we have
	\begin{equation} \label{eqn:scheme5}
		P^{\alpha_X} \Smooth_\delta^{\alpha_X} \bar{f} = P^{\alpha_X} \Smooth_\delta^{\alpha_X} \Ex^{\alpha} f = Q_\delta^{\alpha} f .
	\end{equation}
	Note that instead of trying to determine the actual conditional expectation $\bar{f}$, we approximate a smoothed version of it. This converts the $L^1$ function $\bar{f}$ into a function which is well-defined pointwise and is continuous. Thus $C(X)$ becomes the common space in which both $\Smooth_\delta^{\mu_X} \Ex^{\alpha} f$ and $\Smooth_\delta^{\alpha_X} \Ex^{\alpha} f$ can be compared. This avoids the theoretical challenge of comparing conditional measures, which not only may not converge pointwise, but may not converge in distribution too\citep[e.g.][Cor 4.1]{CrimaldiPratelli2005}. Furthermore, the bandwidth parameter $\delta$ also controls the $L^2$ distance of $\Smooth_\delta^{\mu_X} \Ex^{\alpha} f$ from $\Ex^{\alpha} f$. These two kernels $k$ and $p$ that we use in our theory and algorithm are stated formally below :
	
	\begin{Assumption} \label{A:6}
		\blue{$k$ is a continuous, symmetric, strictly positive definite kernel on $X$, and $p$ is a continuous Markov kernel on $X$.}
	\end{Assumption}
	
	\paragraph{Null hypothesis} A kernel is said to be \emph{cc-universal } \citep[see][]{SriperumbudurEtAl2011} if $\rkhs_{\mu_X}$ is dense in $C(\support(\nu))$, the space of continuous functions on the support of $\nu$.. Kernels such as a $k_{\text{Gauss}}$ are of the form $k(x,y) = \psi(x-y)$, with $\psi$ being a bounded, continuous, integrable function and a Fourier transform of some Borel measure. It has been shown in \cite{SriperumbudurEtAl2011} that such kernels are cc-universal. With this in mind, we assume :
	
	\begin{Assumption} \label{A:3}
		The smoothed conditional expectation $\Smooth^{\mu}_\delta \bar{f}$ 
		lies in $\rkhs_{\mu_X}$.
	\end{Assumption}
	
	Assumption~\ref{A:3} might seem artificial as the choice of the kernel is independent of the unknown function $f$. Thus the function $\bar{f}$ need not lie in $\rkhs_{\mu_X}$. Our first statement in Theorem~\ref{thm:1} does not require Assumption~\ref{A:3}, but uses it to provide a stronger statement. In addition Assumption~\ref{A:3} 
	is not restrictive due to the density of the RKHS in $C(X)$. Given any RKHS $\rkhs$ and an $f\in C(X)$, the sequence of norms
	\[ a_{n} := \inf \SetDef{ \norm{h}_{\rkhs} }{ h\in \rkhs_{\mu_X}, \, \norm{f-h}_{C(X)} < \frac{1}{n} } , \quad n=1,2,\ldots \]
	is called the rate of approximation of $f$. Each of the RKHS approximations can be used as a candidate for $f$ that satisfies Assumption~\ref{A:3}. Its oscillatory nature, captured by $\norm{f}_{\rkhs}$, determines the tuning of the experiment parameters.
	
	Thus, an RKHS supported on the observed data-space $X$ will be our choice of hypothesis space in the estimation of the conditional expectation. An RKHS provides several advantages, it has a Hilbert space structure, pointwise evaluation is a bounded operation, and under mild conditions, they are dense in the space of continuous functions. More importantly, the conditional expectation operator has been shown to be well approximated in operator norm by Hilbert Schmidt operators \cite{MollenhauerKoltai2020}. This gives kernel-based techniques a clear edge over  other techniques. Finally, RKHS in many situations can be endowed with a Banach algebra structure \cite{DasDimitris_CascadeRKHS_2019}, thus enriching them further for Harmonic analysis.
	
	Our main result below is in terms of an \emph{ $\epsilon$-regularized } least squares solution to a linear inverse problem $Ma = b$. This is the solution $a = \paran{ M^T M + \epsilon }^{-1} M^T b $.
	
	\begin{theorem} \label{thm:1}
		Suppose Assumptions~\ref{A:1}, \ref{A:2}, \ref{A:4} and \ref{A:6} hold. Then if $\phi_{\alpha, \nu, \epsilon} \in L^2(\nu)$ is the least-squares solution in $\phi$ to the equation
		\begin{equation} \label{eqn:scheme2}
			P^{\alpha_X} K^\nu \phi = Q_{\delta}^{\alpha} f
		\end{equation}
		then
		\begin{equation} \label{eqn:thm1a}
			\lim_{\alpha \to \mu} \lim_{\epsilon\to 0^+} \norm{ K^\nu \phi_{\alpha, \nu, \epsilon} - \Smooth_{\delta}^{\alpha_X} \bar{f} }_{L^2(\nu)} = 0.
		\end{equation}
		Furthermore, if Assumption \ref{A:3} holds, then
		\begin{equation} \label{eqn:thm1b}
			\lim_{\nu \to \mu_X} \lim_{ \alpha \to \mu , \, \epsilon\to 0^+
			} \lim_{} \norm{ K^\nu \phi_{\alpha, \nu, \epsilon} - \Smooth_{\delta}^{\mu_X} \bar{f} }_{\rkhs} = 0.
		\end{equation}
	\end{theorem}
	
	Theorem~\ref{thm:1} is proved in Section ~\ref{sec:proofs}. 
	
	\paragraph{Remark} Since an RKHS is continuously embedded in $C(X)$, \eqref{eqn:thm1b} implies 
	\begin{equation} \label{eqn:thm1c}
		\lim_{\nu \to \mu_X} \lim_{\alpha \to \mu, \, \epsilon\to 0^+} \norm{ K^\nu \phi_{\alpha, \nu, \epsilon} - \Smooth_\delta^{\alpha_X} \bar{f} }_{C(X)} = 0,
	\end{equation}
	a guarantee of uniform convergence to the $\delta$-smoothed version of the expectation operator.
	
	\paragraph{Remark} A major difference of Claim~\eqref{eqn:thm1b} from Claim~\eqref{eqn:thm1a} is that the former involves a joint limit $\alpha \to \mu$ and $\epsilon\to 0^+$. 
	\blue{The measure $\alpha$ is to be interpreted as the approximation of $\mu$ provided by data. The parameter $\epsilon$ is to be interpreted as the degree of regularization imposed by the learning process. The joint limit in \eqref{eqn:thm1b} indicates that these can be treated independent of the other. This sets Theorem \ref{A:1} apart from most existing techniques in the field.  } 
	
	\paragraph{Remark} Theorem \ref{thm:1} and our subsequent techniques aim to approximate $\Smooth_\delta^{\mu_X} \bar{f}$ instead of $\bar{f}$. As explained in the context of localized kernels, $\bar{f}$ may not be a proper function and does not allow pointwise comparisons. $\Smooth_\delta^{\mu_X} \bar{f}$ can be interpreted as a smoothed or smudged version of $\bar{f}$. The parameter $\delta$ controls the degree of smudging, as quantified in \eqref{eqn:def:local_ker}.
	
	\paragraph{Commutations} The following diagram in \eqref{eqn:scheme3} illustrates the operator theoretic commutations used in our scheme. 
	\begin{equation} \label{eqn:scheme3}
		\begin{tikzcd} [column sep = large]
			& L^2(\nu) & C(X\times Y) \arrow[dashed, PineGreen]{l}[swap]{ T_{\alpha, \nu, \epsilon} } \arrow[dashed, PineGreen, bend left=10]{dl}{ \bar{T}_{\alpha, \nu, \epsilon} } \arrow[blue]{dd}{ \bar{Q}_{\delta}^{\alpha} } \arrow[blue]{r}{ \Ex^{\alpha} }  & L^1(\alpha_X) \arrow[blue]{d}{\subset} \\
			L^2(\nu) \arrow[bend left=20, Choklet]{ur}{ \tilde{K}^\nu } \arrow[Choklet]{r}{K^\nu} & C(X) \arrow[Choklet]{u}{\iota_\nu} & & L^2(\alpha_X) \arrow[blue]{d}{ \Smooth_\delta^{\alpha_X} } \\
			& & L^2(\alpha_X) \arrow[Itranga, bend left=10]{ull}{ B_{\alpha, \nu, \epsilon} } & C(X \arrow[blue]{l}{ \bar{P}^{\alpha_X} }) 
		\end{tikzcd}
	\end{equation}
	The blue loop expresses the identity in \eqref{eqn:scheme5}. The smoothing operator $Q^{\alpha_X}$ is shown as the composite of the conditional expectation operator, and the smoothing operator $P^{\alpha_X}$. The map $B_{\alpha, \nu, \epsilon}$ shown in red is the linear map that provides the $\epsilon$-regularized least squares solution to \eqref{eqn:scheme2}. It is explicitly constructed later in \eqref{eqn:def:B_oprtr}, and explored in more detail in Section~\ref{sec:proofs}. The commutation shown in brown is the action of the smoothing operator on the result of the linear inverse problem. As a result, $T_{\alpha, \nu, \epsilon}, \bar{T}_{\alpha, \nu, \epsilon}$ are respectively the $L^2(\alpha_X)$ and continuous versions of the estimation technique.
	
	The various colored paths represent not only the mathematical aspect of commutation loops, but also the practical aspects of the technique. In a data-driven application, both $\alpha, \nu$ are sampling measures built from data. The Markov operator $\bar{Q}_{\delta}^{\alpha}$ then behaves similarly as a moving average, and helps overcome the scarcity of samples along individual leaves of the partition $\SetDef{ \{x\}\times Y }{ x\in X }$. The commutation ensures that although $\bar{Q}_{\delta}^{\alpha}$ is easily constructible from samples of the function $f\in C(X\times Y)$, it bears a meaningful relation with the conditional expectation. The green loop represents the difference between in-sample and out-of sample extensions. Although $\tilde{K}^\nu$ and $K^\nu$ are related by the simple inclusion map $\iota_\nu$, they have different implementations. Although $\tilde{K}^\nu$ is expressed as the composition of $K^\nu$ with $\iota_\nu$, it has a more direct and immediate evaluation.
	
	We next prove Theorem~\ref{thm:1}, by taking a closer look at the operators and spaces in the background.
	
	
	\section{A closer look at Theorem~\ref{thm:1}} \label{sec:proofs}
	
	We prove Theorem~\ref{thm:1} in this section, and begin by being more particular with our notation. In the various diagrams below, we use dashed, green arrows to indicate that a new operator is being defined via construction. Given any continuous kernel $k$ and a finite measure $\nu$ supported on $X$, one has the following diagram of spaces and operators :
	\[\begin{tikzcd}
		& C(X) \arrow{r}{\iota_\nu} & L^2(\nu) \\
		C(X) \arrow[bend left=10, dashed, PineGreen]{ur}{\bar{K}^\nu} \arrow{r}{\iota_\nu} & L^2(\nu) \arrow{u}{K^\nu} \arrow[bend right=10, dashed, PineGreen]{ur}[swap]{\tilde{K}^\nu}
	\end{tikzcd}\]
	Here $\iota_\nu : C(X) \to L^2(\nu)$ is the inclusion of continuous maps into the space of square integrable maps. The operators $\bar{K}^\nu$ and $\tilde{K}^\nu$ are respectively pre and post compositions of $K^\nu$ with $\iota_\nu$. We use the analogous notation for $P, \tilde{P}$ and $\bar{P}$ too. With this notation in mind, \eqref{eqn:scheme3} will be rewritten as 
	\begin{equation} \label{eqn:scheme8}
		\begin{tikzcd} [column sep = large]
			& & L^1(\alpha_X) \arrow{r}{\subset} & L^2(\alpha_X) \arrow{d}{ \Smooth_\delta^{\alpha_X} } \\
			& L^2(\nu) & C(X\times Y) \arrow[dashed, PineGreen]{r}{ \Ex^{\alpha, \delta} } \arrow{u}{ \Ex^{\alpha} } \arrow[dashed]{l}[swap]{ T_{\alpha, \nu, \epsilon} } \arrow[dashed, bend left=10]{dl}{ \bar{T}_{\alpha, \nu, \epsilon} } \arrow{dd}{ \bar{Q}_{\delta}^{\alpha} } & C(X ) \arrow{d}{\iota_\alpha} \\
			L^2(\nu) \arrow[bend left=20]{ur}{ \tilde{K}^\nu } \arrow{r}{K^\nu} & C(X) \arrow{u}{\iota_\nu} & & L^2(\alpha_X) \arrow{dl}{ \tilde{P}^{\alpha_X} } \\
			& & L^2(\alpha_X) \arrow[bend left=10]{ull}{ B_{\alpha, \nu, \epsilon} }
		\end{tikzcd}
	\end{equation}
	We have named the composite operator $ \Smooth_\delta^{\alpha_X} \Ex^{\alpha}$ as $ \Ex^{\alpha, \delta}$. \blue{Our idea of proving Theorem \ref{thm:1}  depends on factoring the various kernel integral operators on both sides of \eqref{eqn:scheme2} into a composition of simpler operators, We also intend to cast the entire procedure of finding the $\epsilon$-regularized least squares solution into an operator too. The diagram in \eqref{eqn:scheme8} will be the first of a sequence of expansions of the commutative diagram in \eqref{eqn:scheme3}, which will reveal these decomposition.}
	
	\blue{First we examine the left hand side (LHS) of \eqref{eqn:scheme3}.} Let $P^{\alpha_X}$ and $K^\nu$ respectively be the kernel integral operators corresponding to the Markov kernel $p$ and symmetric s.p.d. kernel $k$, and probability measures $\alpha$ and $\nu$. This leads to 
	\begin{equation} \label{eqn:def:A_oprtr}
		\begin{tikzcd}
			& L^2(\alpha_X) & C(X) \arrow[l, "\iota_\alpha"'] \\
			L^2(\nu) \arrow[r, "K^\nu"'] \arrow[dashed, PineGreen]{ru}{ A_{\alpha, \nu} } & C(X) \arrow[r, "\iota_\alpha"'] & L^2(\alpha_X) \arrow[bend right=10]{lu}{ \tilde{P}^{\alpha_X} } \arrow[u, "P^{\alpha_X}"']
		\end{tikzcd}
	\end{equation}
	The operator $A_{\alpha, \nu}$ so constructed is effectively the LHS of \eqref{eqn:scheme2}. \blue{Next we determine the operator form of $B_{\alpha, \nu,\epsilon}$}, the $\epsilon$-regularized pseudo-inverse of $A_{\alpha, \nu}$ :
	\begin{equation} \label{eqn:def:B_oprtr}
		\begin{tikzcd}
			& L^2(\nu) \\
			L^2(\alpha_X) \arrow[r, "{A_{\nu}^*}"'] \arrow[PineGreen, dashed]{ur}{B_{\alpha, \nu,\epsilon}} & L^2(\nu) \arrow{u}[swap]{ \paran{ A_{\alpha, \nu}^* A_{\alpha, \nu} + \epsilon }^{-1} }
		\end{tikzcd}
	\end{equation}
	The operators $A_{\alpha, \nu}$ and $B_{\alpha, \nu,\epsilon}$ are the building blocks to constructing the solution to \eqref{eqn:scheme2}, as shown in \eqref{eqn:scheme3}. \blue{The third and final operator theoretic realization is that of Assumption \ref{A:4}.} The condition of absolutely continuity implies the following commuting diagram
	\begin{equation} \label{eqn:A4}
		\begin{tikzcd}
			C(X) \arrow{d}[swap]{\iota_\alpha} \arrow{r}{\iota_\nu} & L^2(\nu) \arrow[dashed, PineGreen]{dl}{ j_{\nu\to\alpha} } \\
			L^2(\alpha_X)
		\end{tikzcd}
	\end{equation}
	The map $j_{\nu\to\alpha}$ is a simple inclusion map, and is built on the fact that any function which is $L^2(\nu)$ integrable is also $L^2(\alpha_X)$ integrable. 
	The simple commutation in \eqref{eqn:A4} has several important consequences throughout our proof. The first is 
	
	\begin{lemma} \label{lem:ue93}
		The operator $A_{\alpha, \nu}$ is a compact and injective operator.
	\end{lemma}
	
	Lemma~\ref{lem:ue93} is proved in Section~\ref{sec:proof:ue93}. To get more out of \eqref{eqn:A4}, we expand it to 
	\[\begin{tikzcd}
		C(X) \arrow{d}[swap]{\iota_\alpha} \arrow{r}{\iota_\nu} & L^2(\nu) \arrow{dl}{ j_{\nu\to\alpha} } \arrow[bend left=30]{r}{ \paran{ \tilde{K}^\nu }^{-1} } & L^2(\nu) \arrow[bend left=30]{l}{ \tilde{K}^\nu } \\
		L^2(\alpha_X) 
	\end{tikzcd}\]
	and then
	\[\begin{tikzcd}
		C(X) \arrow{d}[swap]{\iota_\alpha} \arrow{r}{\iota_\nu} & L^2(\nu) \arrow{dl}{ j_{\nu\to\alpha} } \arrow[bend left=30]{r}{ \paran{ \tilde{K}^\nu }^{-1} } & L^2(\nu) \arrow[bend left=30]{l}{ \tilde{K}^\nu } \arrow{d}{K^{\nu}} \\
		L^2(\alpha_X) & & C(X) \arrow{ll}{\iota_{\alpha}}
	\end{tikzcd}\]
	This commutations added to \eqref{eqn:scheme8} gives :
	\begin{equation} \label{eqn:scheme4}
		\begin{tikzcd} [column sep = large]
		& L^1(\alpha_X) \arrow{r}{\subset} & L^2(\alpha_X) \arrow{d}{ \Smooth_\delta^{\alpha_X} } \\
		L^2(\nu) & C(X\times Y) \arrow{r}{ \Ex^{\alpha, \delta} } \arrow{l}[swap]{ T_{\alpha, \nu, \epsilon} } \arrow[bend left=10]{dl}{ \bar{T}_{\alpha, \nu, \epsilon} } \arrow{dd}{ \tilde{Q}^\alpha } \arrow{u}{ \Ex^{\alpha} }  & C\paran{ X } \arrow{d}[swap]{\iota_\alpha} \arrow{r}{\iota_\nu} & L^2(\nu) \arrow[bend left=80]{dd}{ \paran{ \tilde{K}^\nu }^{-1} } \arrow{dl}{ j_{\nu\to\alpha} } \\
		C(X) \arrow{u}{\iota_\nu} & & L^2(\alpha_X) \arrow{dl}[swap]{ \tilde{P}^{\alpha_X} } \\
		L^2(\nu) \arrow[bend left=80]{uu}{ \tilde{K}^\nu } \arrow{u}{K^\nu} & L^2(\alpha_X) \arrow{l}{ B_{\alpha, \nu, \epsilon} } & C(X) \arrow{u}{ \iota_\alpha } & L^2(\nu) \arrow{l}{ K^\nu } \arrow{uu}[swap]{ \tilde{K}^\nu } \arrow[bend left=30]{ll}{ A_{\alpha, \nu} }
	\end{tikzcd}
	\end{equation}
	This allows us to write
	\[\tilde{Q}^{\alpha_X} = \tilde{P}^{\alpha_X} \iota_\alpha K^\nu \paran{ \tilde{K}^\nu }^{-1} \iota_\nu \Ex^{\alpha, \delta} = A_{\alpha, \nu} \paran{ \tilde{K}^\nu }^{-1} \iota_\nu \Ex^{\alpha, \delta} ,\]
	where the last equality follows from \eqref{eqn:def:A_oprtr}. Applying $\tilde{K}^\nu B_{\alpha, \nu, \epsilon}$ on both sides gives :
	\begin{equation} \label{eqn:ddo30}
		\tilde{K}^\nu B_{\alpha, \nu, \epsilon} \bar{Q}_{\delta}^{\alpha} = \tilde{K}^\nu B_{\alpha, \nu, \epsilon} A_{\alpha, \nu} \paran{ \tilde{K}^\nu }^{-1} \iota_\nu \Ex^{\alpha, \delta} .
	\end{equation}
	%
	
	\begin{lemma} \label{lem:inr0}
		The operator $B_{\alpha, \nu, \epsilon}$ is compact. Moreover, the following limit holds pointwise :
		%
		\[ \lim_{\epsilon\to 0^+} B_{\alpha, \nu, \epsilon} A_{\alpha, \nu} = \Id_{ L^2(\nu) } . \]		%
	\end{lemma}
	
	Lemma~\ref{lem:inr0} is proved in Section~\ref{sec:proof:inr0}. 
	Using Lemma~\ref{lem:inr0} one can establish the limits of \eqref{eqn:ddo30} :
	\begin{equation} \label{eqn:scheme7}
		\lim_{\epsilon\to 0^+} \tilde{K}^\nu B_{\alpha, \nu, \epsilon} \bar{Q}_{\delta}^{\alpha} f = \tilde{K}^\nu \paran{ \tilde{K}^\nu }^{-1} \iota_\nu \Ex^{\alpha, \delta} f = \iota_\nu \Ex^{\alpha, \delta} f .
	\end{equation}
	To study the limit of \eqref{eqn:scheme4}, we use the following lemma.
	
	\begin{lemma} \label{lem:do3}
		For every $f\in C(X; C(Y))$, $\Ex^{\alpha, \delta} f$ converges uniformly to $\Ex^{\mu, \delta} f$ as $\alpha$ converges weakly to $\mu$.
	\end{lemma}
	
	Lemma~\ref{lem:do3} is proved in Section \ref{sec:proof:do3}. \blue{ This completes the preparatory phase of the proof. }
	
	\paragraph{Proof of \eqref{eqn:thm1a}} Lemma~\ref{lem:do3} applied to \eqref{eqn:scheme4} gives :
	\begin{equation} \label{eqn:scheme6}
		\begin{split}
			\lim_{\alpha\to\mu} \lim_{\epsilon\to 0^+} \tilde{K}^\nu B_{\alpha, \nu, \epsilon} \bar{Q}_{\delta}^{\alpha} f &= \lim_{\alpha\to\mu} \iota_\nu \Ex^{\alpha, \delta} f , \quad \mbox{ by \eqref{eqn:scheme4}} \\
			&= \iota_\nu \lim_{\alpha\to\mu} \Ex^{\alpha, \delta} f \\
			& = \iota_\nu \Ex^{\mu, \delta} f , \quad \mbox{ by Lemma~\ref{lem:do3}} .
		\end{split}
	\end{equation}
	Equation \eqref{eqn:scheme6} states that if the least squares solution 
	\[ \phi_{\alpha, \nu, \epsilon} := B_{\alpha, \nu, \epsilon} \bar{Q}_{\delta}^{\alpha} f , \]
	is smoothed using $K^\nu$, then the resulting function converges in $L^2(\nu)$ norm to the conditional expectation. 
	
	\paragraph{Proof of \eqref{eqn:thm1b}}  To proceed with the next part of the theorem, we reuse the notation $K^\nu$ to also denote the map of $L^2(\nu)$ into $\rkhs_\nu$. Thus we have the following commutation of maps
	\[\begin{tikzcd}
		L^2(\nu) \arrow{dr}[swap]{ K^\nu } \arrow{r}{ K^\nu } & C(X) \\
		& \rkhs_\nu \arrow{u}[swap]{ \subset }
	\end{tikzcd}\]
	Since $\support( \nu ) \subseteq \support( 1\mu_X )$, the space $\rkhs_\nu$ is a subspace of $\rkhs_{\mu_X}$. Thus there is a projection between these spaces. The following commutation provides an alternate interpretation of this projection.
	\begin{equation} \label{eqn:pi4o4}
		\begin{tikzcd}
			\rkhs_{\mu_X} \arrow{dr}[swap]{ \proj } \arrow{r}{ \subset } & C(X) \arrow{r}{ \iota_\nu } & L^2(\nu) \arrow{d}{ \paran{ \tilde{K}^\nu }^{-1} } \\
			& \rkhs_{\nu} & L^2(\nu) \arrow{l}{ K^\nu }
		\end{tikzcd}
	\end{equation}
	Next we state another important consequence of the hypothesis in Assumption \ref{A:3}.
	
	\begin{lemma} \label{lem:kpod9}
		Suppose Assumptions~\ref{A:1}, \ref{A:2}, \ref{A:4} and \ref{A:3} holds. Then 
		\[ \lim_{\alpha\to \mu, \epsilon\to 0^+} \norm{ \paran{ B_{\alpha, \nu, \epsilon} A_{\alpha, \nu} - \Id } \paran{ \tilde{K}^\nu }^{-1} \iota_\nu \Ex^{\mu, \delta} f }_{L^2(\nu)} < \infty . \]
	\end{lemma}
	
	Lemma~\ref{lem:kpod9} is proved in Section \ref{sec:prof::kpod9}. This leads to 
	\[\begin{split}
		K^\nu B_{\alpha, \nu, \epsilon} \bar{Q}_{\delta}^{\alpha} f & = K^\nu B_{\alpha, \nu, \epsilon} A_{\alpha, \nu} \paran{ \tilde{K}^\nu }^{-1} \iota_\nu \Ex^{\alpha, \delta} f , \quad \mbox{ by \eqref{eqn:ddo30} } \\
		& \xrightarrow{\epsilon\to 0^+ , \, \alpha \to \mu} K^\nu \paran{ \tilde{K}^\nu }^{-1} \iota_\nu \Ex^{\alpha, \delta} f, \quad \mbox{by Lemma~\ref{lem:kpod9} } \\
		& \xrightarrow{\epsilon\to 0^+ , \, \alpha \to \mu} K^\nu \paran{ \tilde{K}^\nu }^{-1} \iota_\nu \Ex^{\alpha, \delta} f, \quad \mbox{by Lemma~\ref{lem:do3} } \\
		& = \proj_{ \rkhs_\nu } \Ex^{\mu} f , \quad \mbox{ by \eqref{eqn:pi4o4} }  \\
		&	 \xrightarrow{ \nu \to \mu_X  } \Ex^{\mu} f.
	\end{split}\]
	Thus $K^\nu \phi_{\alpha, \nu, \epsilon}$ converges to $g$ in RKHS norm. 	
	This completes the proof of Theorem~\ref{thm:1}. \qed
	
	In the next section we look at a practical implementation of this scheme, and the accompanying guarantee of convergence.
	
	\section{Numerical implementation} \label{sec:numeric}
	
	In a data-driven implementation, the inputs to any numerical recipe is a dataset, along with some algorithmic parameters. We assume that the data originates as follows :
	
	\begin{Assumption} \label{A:5}
		There is a sequence of points $(x_n, y_n) \in X\times Y$ for $n=1,2,3,\ldots$, equidistributed with respect to the probability measure $\mu$ from Assumption~\ref{A:1}.
	\end{Assumption}
	
	The concept of equidistribution is a major relaxation of the assumption of being i.i.d.. Such an assumption has been utilized with great success in the theoretical understanding of numerical methods for timeseries which have strong correlations, such as those arising from dynamical systems \citep[e.g.][]{DasGiannakis_delay_2019, DGJ_compactV_2018, DasGiannakis_RKHS_2018}. Algorithm~\ref{algo:1} below presents our main procedure.
	
	\begin{algo} \label{algo:1}
		RKHS representation of conditional expectation.
		\begin{itemize}
			\item \textbf{Input.} A sequence of pairs $\SetDef{(x_n, y_n)}{ n=1,\ldots,N}$ with $x_n\in\real^d$ and $y_n\in \real$.
			\item \textbf{Parameters.}
			\begin{enumerate}
				\item Choice of RKHS kernel $k:\real^d \times \real^d \to \real^+$.
				\item Smoothing parameter $\delta>0$.
				\item Sub-sampling parameter $M\in\num$ with $M<N$.
				\item Regularization parameter $\epsilon$.
			\end{enumerate}
			\item \textbf{Output.} A vector $\vec{a} = \paran{a_1, \ldots , a_M} \in \real^M$ such that
			\[ (\Ex^{\alpha} f)(x) \approx \sum_{m=1}^{N} a_m k(x, x_m) , \quad \forall x \in \real^d . \]
			\item \textbf{Steps.}
			\begin{enumerate}
				\item Compute a Gaussian Markov kernel matrix using \eqref{eqn:def:GaussSymm} :
				\begin{equation} \label{eqn:GM}
					\Matrix{G_{\delta}} \in \real^{N\times N}, \quad \Matrix{G_\delta}_{i,j} = k_{ \text{Gauss}, \delta }^{ \text{symm}, \beta } (x_i, x_j) .
				\end{equation}
				\item Compute a Markov kernel $\Matrix{P} \in \real^{N\times N}$ as $\Matrix{P}_{i,j} := p(x_i, x_j)$
				\item Compute the kernel matrix $\Matrix{K}\in \real^{N\times M}$ as  $\Matrix{K}_{i,j} = k(x_i, x_j)$.
				\item Find a vector $\vec{a} \in \real^M$ as the $\epsilon$-regularized least-squares solution to the equation
				\[\Matrix{P} \Matrix{K} \vec{a} = \Matrix{P} \Matrix{G_\delta} \vec{y} . \]
			\end{enumerate}
		\end{itemize}
	\end{algo}
	
	Algorithm~\ref{algo:1} has two components, the choice of an RKHS kernel, and the creation of a Markov kernel which approximates  the smoothing operator. We usually choose $p$ to be the Markov normalized Gaussian kernel from \eqref{eqn:def:GaussSymm} ,
	\[p(z,x) := \exp\paran{ - \dist \paran{z-x}^2/\delta } / \int_X \exp\paran{ -\dist \paran{z-y}^2/\delta } d\mu_X(y) .\]
	Theorem~\ref{thm:2} below provides an interpretation of the output vector $\vec{a}$ from Algorithm~\ref{algo:1}, and the nature of the convergence of the results.
	
	\begin{theorem} \label{thm:2}
		Suppose Assumptions~\ref{A:1}, \ref{A:2}, \ref{A:6}, \ref{A:3} and \ref{A:5} hold. Let $\vec{a}$ be the output of Algorithm~\ref{algo:1} is applied to the data $\paran{ x_n, y_n }_{n=1}^{N}$. Then 
		\begin{equation} \label{eqn:thm2}
			\lim_{M\to \infty} \lim_{N\to\infty, \epsilon \to 0^+} \norm{ \sum_{n=1}^{N} a_n k(\cdot, x_n) - \Smooth_\delta^\mu \bar{f} }_{\rkhs} = 0.
		\end{equation}
	\end{theorem}
	
	Note that Theorem~\ref{thm:2} is independent of the choice of the kernel $k$ in Algorithm~\ref{algo:1}. Algorithm~\ref{algo:1} itself can be carried out on any dataset $(x_n,y_n)$, irrespective of whether any of Assumptions~\ref{A:1}, \ref{A:2}, \ref{A:3} and \ref{A:5} hold. Assumptions~\ref{A:1} and \ref{A:5} are needed to place the dataset in context, and Assumptions \ref{A:2} and \ref{A:3} are required to guarantee their convergence. Theorem~\ref{thm:2} is proved in Section~\ref{sec:proof:2}, and is a direct consequence of Theorem~\ref{thm:1}. 
	
	We next apply Algorithm~\ref{algo:1} to a few practical problems.
	
	\section{Examples} \label{sec:example}
	
	Our choice of kernel in all the experiments is the \emph{diffusion} kernel  \cite[e.g.][]{MarshallCoifman2019, WormellReich2021}. Among its various constructions, we choose the following :
	\begin{equation} \label{eqn:def:kdiff}
		\begin{split}
			& k_{ \text{diff}, \epsilon}^\mu (x,y) = \frac{ k_{\text{Gauss}, \epsilon}(x,y) }{ \deg_l(x) \deg_r(y) } ,\\ 
			& \deg_r(x) := \int_X k_{\text{Gauss}, \epsilon}(x,y) d\mu(y) , \quad \deg_l(x) := \int_X k_{\text{Gauss}, \epsilon}(x,y) \frac{1}{ \deg_r(x) } d\mu(y) .
		\end{split}
	\end{equation}
	Diffusion kernels have been shown to be good approximants of the local geometry in various different situations \citep[e.g.][]{BDGV_spectral_2020, CoifmanLafon2006, HeinEtAl2005, VaughnBerryAntil2019}, and are a natural choice for non-parametric learning. It has the added advantage of being symmetrizable :
	\begin{equation} \label{eqn:symm}
		\rho(x) k_{ \text{diff}, \epsilon}^\mu (x,y) \rho(y)^{-1} = \tilde{k}_{ \text{diff}, \epsilon}^\mu (x,y) = \frac{k_{\text{Gauss}, \epsilon}(x,y)}{ \SqBrack{ \deg_r(x) \deg_r(y) \deg_l(x) \deg_l(y) }^{1/2} }, 
	\end{equation}
	where
	\[ \rho(z) = \deg_l(z)^{1/2} / \deg_r(z)^{1/2} . \]

	The kernel $\tilde{k}_{ \text{diff}, \epsilon}^\mu$ from \eqref{eqn:symm} is clearly symmetric. Since it is built from the s.p.d. kernel $k_{ \text{Gauss}, \epsilon}$, $\tilde{k}_{ \text{diff}, \epsilon}^\mu$ is s.p.d. too and thus generates an RKHS of its own. Moreover, the kernel $k_{ \text{diff}, \epsilon}^\mu$  can  be symmetrized by a degree function $\rho$, which is both bounded and bounded above $0$. Such a kernel will be called \emph{RKHS-like}. Let $M_\rho$ be the multiplication operator with $\rho$. Then
	\[ \ran K^\mu_{ \text{diff}, \epsilon} = \ran M_\rho \circ \tilde{K}^\mu_{ \text{diff}, \epsilon} . \]
	Again, because of the properties of $\rho$, both $M_\rho$ and its inverse are bounded operators. Thus there is a bijection between the RKHS generated by $\tilde{k}_{ \text{diff}, \epsilon}^\mu$, and the range of the integral operator $K_{ \text{diff}, \epsilon}^\mu$.
	
	\subsection{Denoising} \label{sec:img_denoise}
	
\begin{figure}[!ht]\center
		\includegraphics[width=.95\linewidth]{\figs 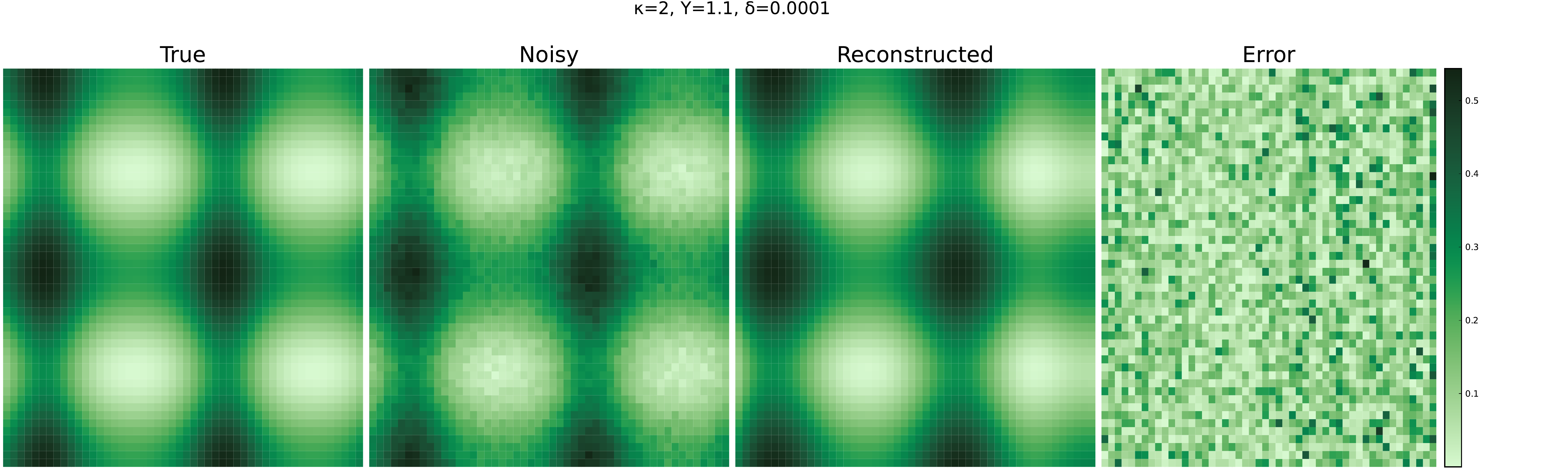}
		
		\includegraphics[width=.95\linewidth]{\figs 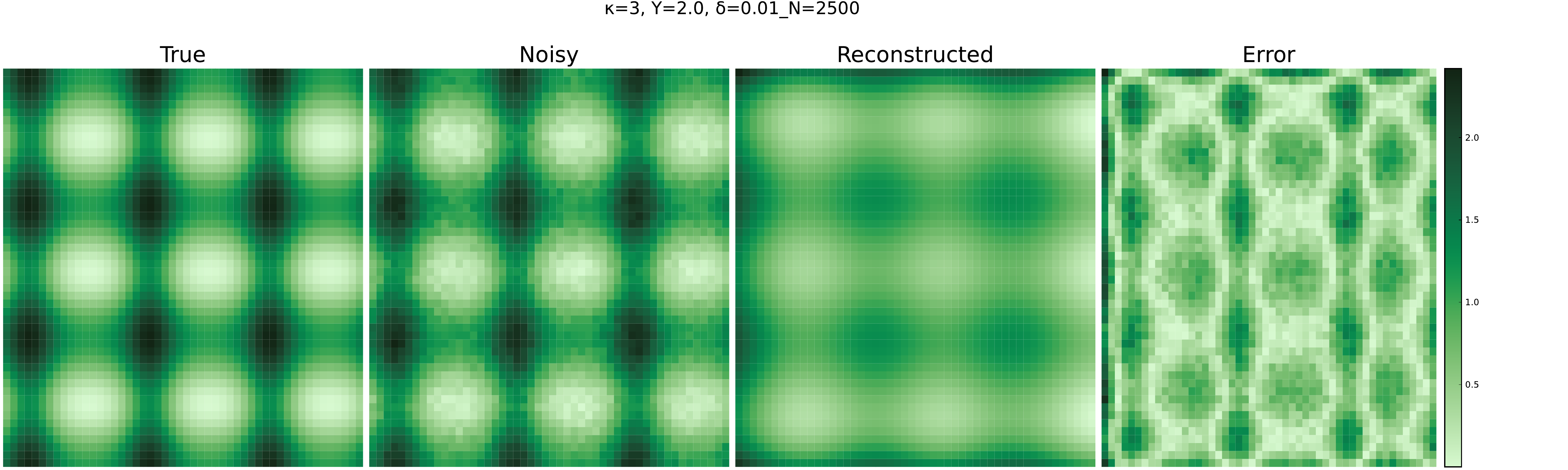}
		
		\includegraphics[width=.95\linewidth]{\figs 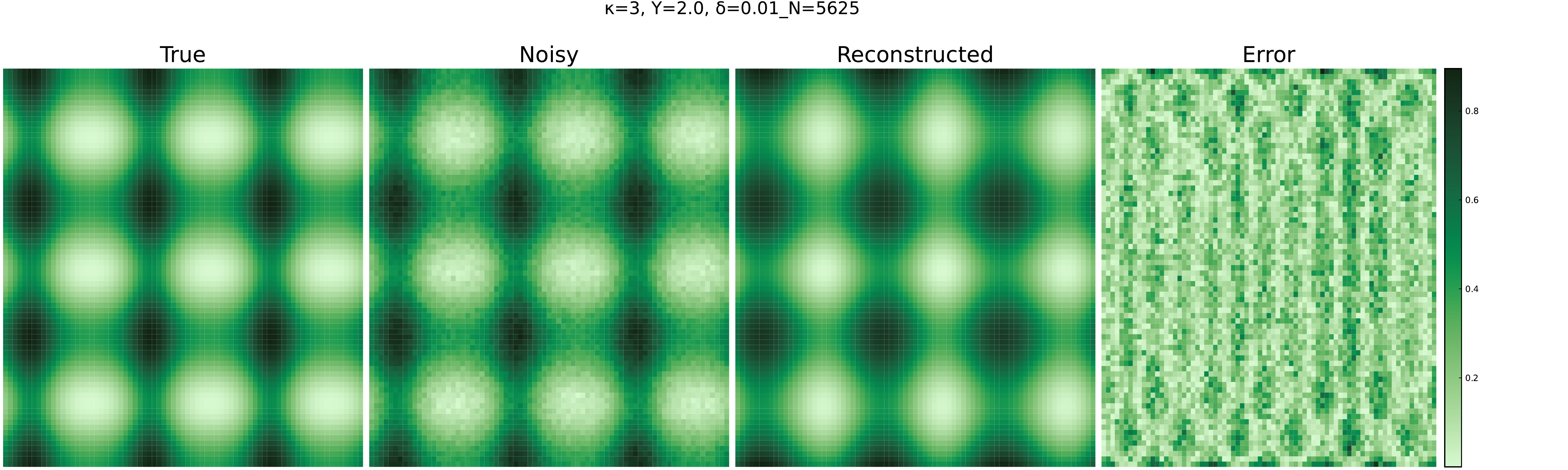}
		\caption{Denoising a monochromatic image. Such an image can be expressed as a continuous function of x--y coordinates. The mathematical formulation is the problem is described in Section~\ref{sec:img_denoise}. The test-image shown here is described by \eqref{eqn:img_denoise}. The parameter $\kappa$ is an index of the $C^1$ norm of the function. The first row shows that Algorithm~\ref{algo:1} performs reasonably well for $\kappa=2$ on a $50\times 50$ pixel image, but the performance deteriorates when $\kappa=2$. The third row shows a much improved result when the image gets more detailed with an increased size of $75\times 75$. }
		\label{fig:as3d}
	\end{figure}

	As explained in Section~\ref{sec:intro}, denoising is a particular instance of Assumptions \ref{A:1} and \ref{A:2}. We illustrate an application of Algorithm~\ref{algo:1} to continuous images in Figures \ref{fig:as3d}. The task of imaging discontinuous images involves many other considerations such as edge detection, and is postponed to a later study. Each of the RGB components of a continuous image may be considered to be points on the graph of a continuous map $\bar{f} : [0,1]^2 \to \real$. The points correspond to the image under $\bar{f}$ of points on a rectangular lattice within $[0,1]^2$. The noise can be considered as an addition from a Gaussian random variable drawn from $\real$. We chose for $\bar{f}$ the function
	\begin{equation} \label{eqn:img_denoise}
		\bar{f} : [0,1]^2 \to \real, \quad \bar{f}(x_1, x_2) := \cos \paran{ \kappa 2\pi x } + e^{ \sin \paran{ \kappa 2 \pi y } }.
	\end{equation}
	where $\kappa\in\num$ is an index for the $C^1$ norm of the image. 	Theorem~\ref{thm:2} states that the convergence or accuracy of the results are dependent on increasing the number of data samples. This presents as a problem in image denoising, as the number of data samples is exactly the number of pixels, and is usually fixed and limited. As a result, the outcome of the numerical procedure becomes sensitive to the smoothing parameter parameter $\delta$ and the $C^1$ norm of the true image.
	
	\subsection{Principal curves - electrostatic charge} \label{sec:principal}
	
\begin{figure}[!ht]\center
		\includegraphics[width=.45\linewidth]{\figs 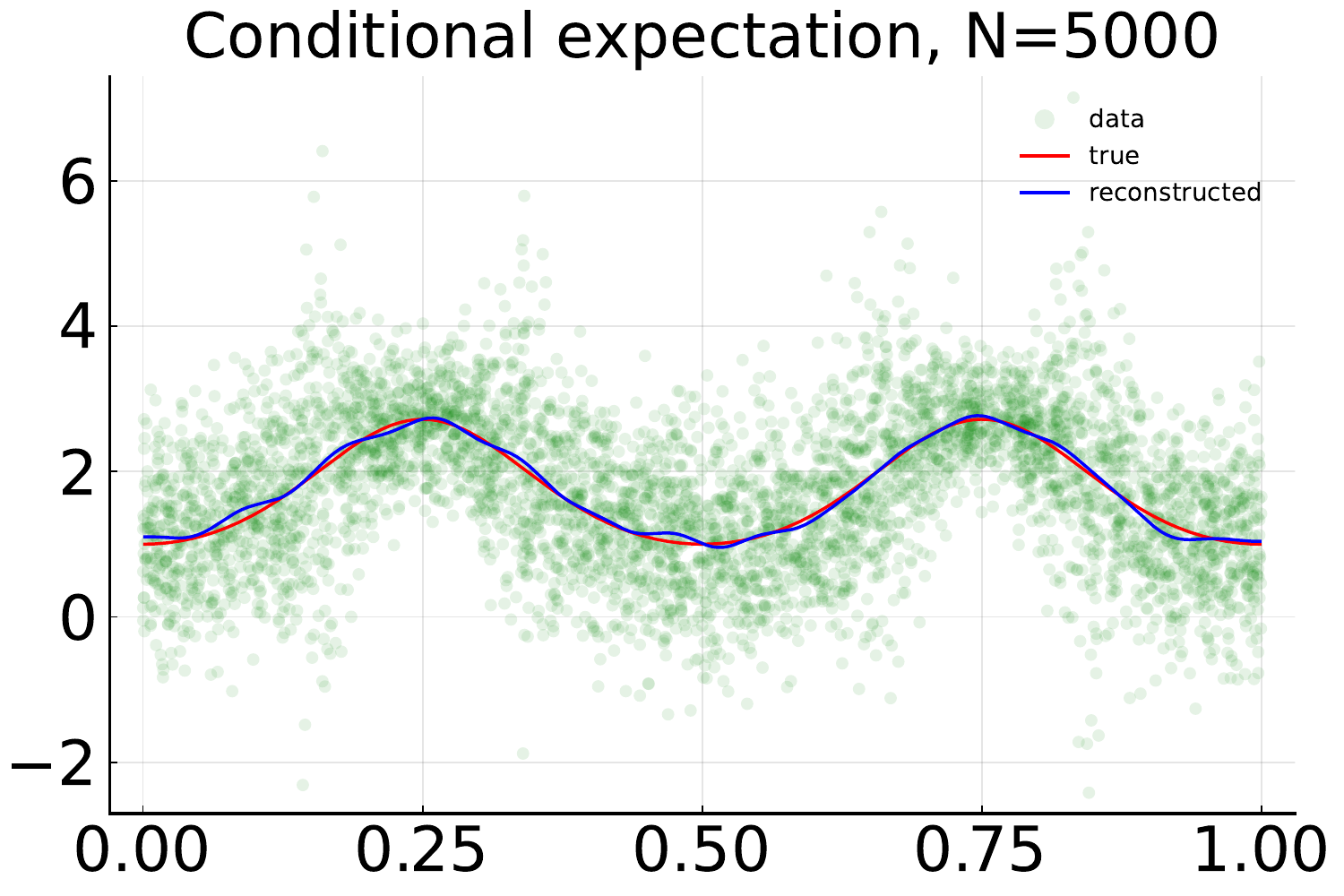}
		\includegraphics[width=.45\linewidth]{\figs 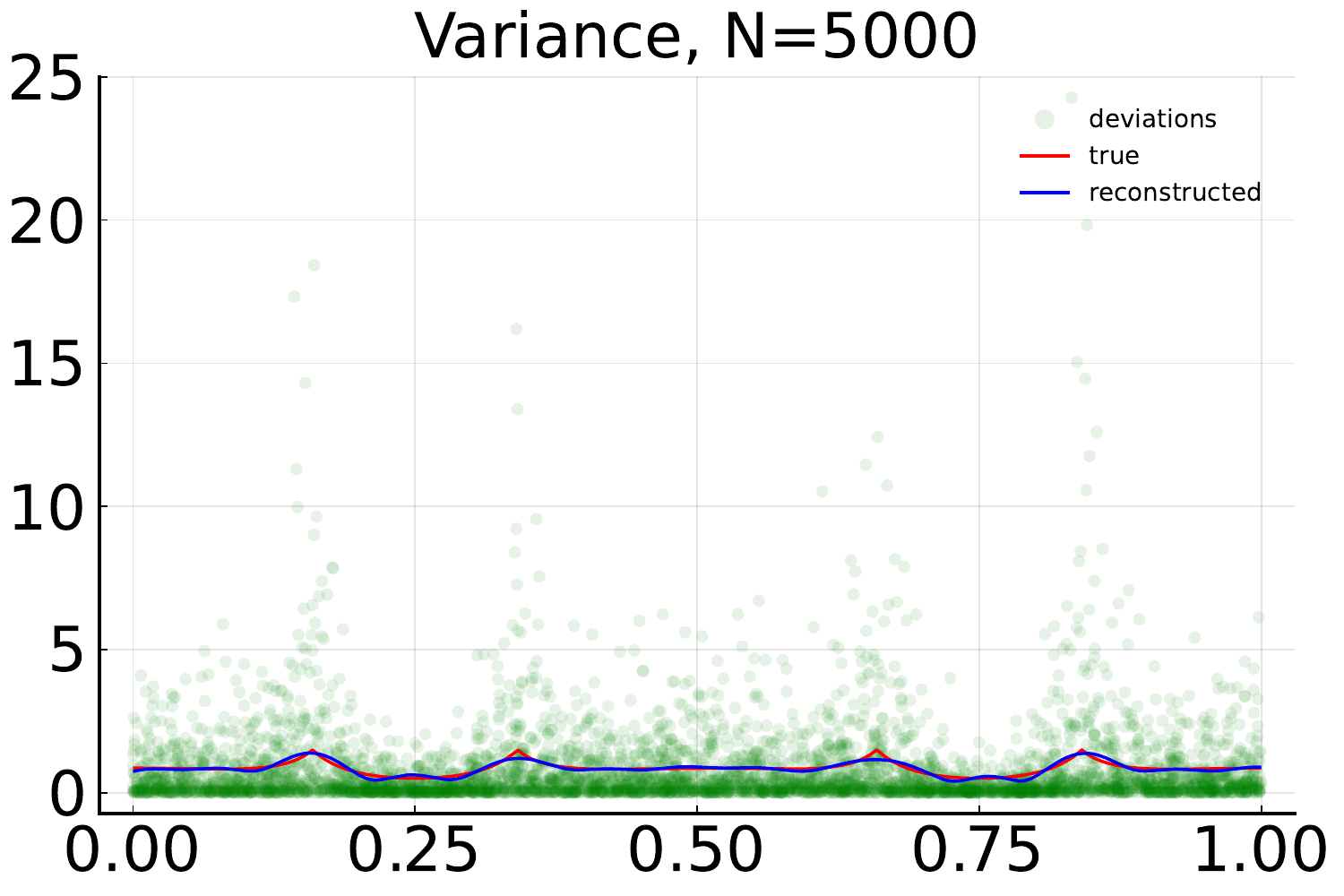}
		
		\includegraphics[width=.45\linewidth]{\figs 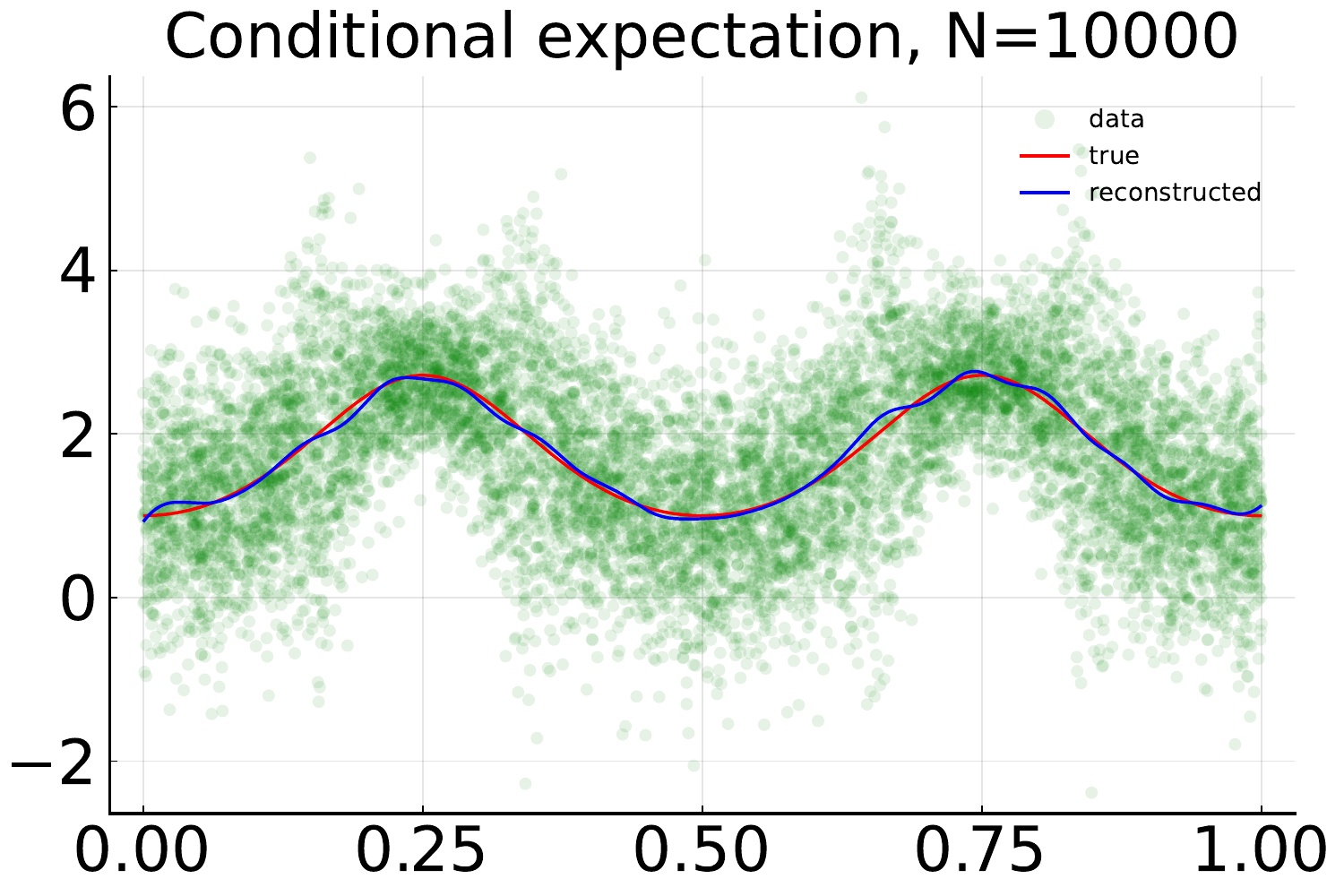}
		\includegraphics[width=.45\linewidth]{\figs 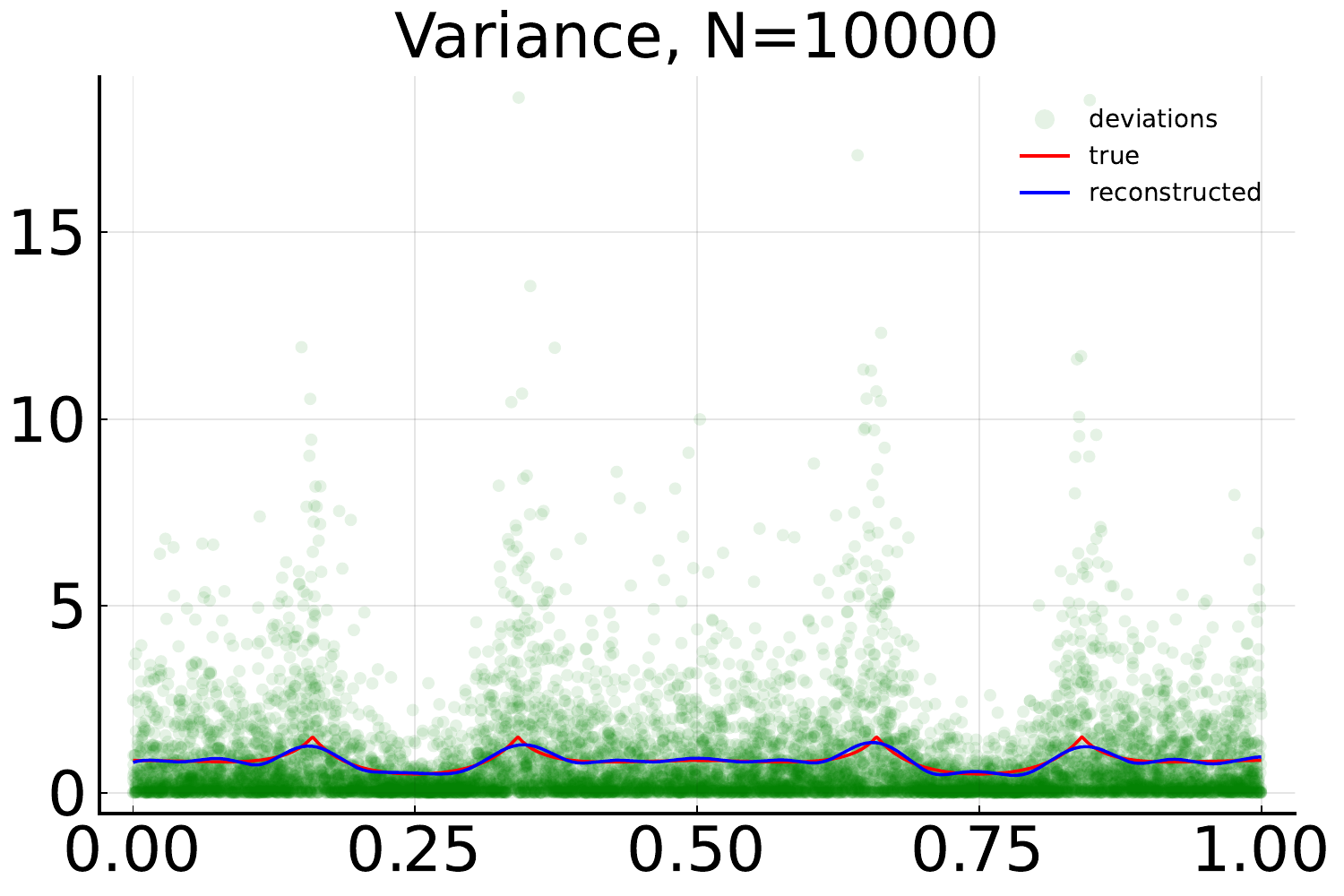}
		\caption{Principal curve estimation. Section~\ref{sec:principal} presents an example of a principal curve problem, from data-points scattered around a "true" or "principal" curev. Equation \eqref{eqn:elctrc2} is a realization of Assumptions \ref{A:1} and \ref{A:2}, and presents a simplified view of electrostatic charge distribution along a wire. We assume that the function $\lambda$ takes the form in \eqref{eqn:elctrc1}. The left panels above show the results of applying Algorithm~\ref{algo:1} to data equidistributed with respect to this distribution, to recover the conditional expectation as a function over $X=[0,1]$. The results show a close match with the true mean, which is simply the curve $\lambda$. The results also visibly improve as the number of samples are increased. The right panel shows a repeated use of Algorithm~\ref{algo:1} to reconstruct the variance as a function over $X=[0,1]$. Again, the results show a strong match with the true function, which is $\rho$. }
		\label{fig:principal}
	\end{figure}

	Given any $C^2$ curve $\lambda : [0,1] \to \real^+$, one can define the function
	\begin{equation} \label{eqn:elctrc2}
		f : [0,1] \times \real \to \real, \quad (x, y) \mapsto \lambda(x) + y / \rho(x), \quad \rho(x) := \frac{3}{ 2 + C \abs{ \lambda''(x) } } , \quad C:= 4 / \norm{\lambda''}_{\sup} .
	\end{equation}
	%
	Equation~\eqref{eqn:elctrc2} is a simplified model of electrostatic charged distribution on curved surfaces. The function $\rho$ controls the spread or variance of ponits around the mean value. By design, $\rho$ has a range in $[0.5,1.5]$. For our test case, we choose for $\lambda$ the function
	\begin{equation} \label{eqn:elctrc1}
		\lambda(x) = \exp \paran{ \sin(2\pi x)^2 }, \quad \forall x\in [0,1] .
	\end{equation}
	The two derivatives of $\lambda$ are :
	\[\begin{split}
		\lambda'(x) &= 2 \pi \lambda(x) \sin(4\pi x) \\
		\lambda''(x) &= 8 \pi^2 \lambda(x) \cos(4\pi x) + 4\pi^2 \lambda(x) \sin(4\pi x)^2 \\
		&= 4 \pi^2 \lambda(x) \SqBrack{ 2 \cos(4\pi x) + \sin(4\pi x)^2 } \\
		&= 4\pi^2 \lambda(x) \SqBrack{ 2 -\SqBrack{\cos(4\pi x)-1}^2  } .
	\end{split}\]
	Let us assign spaces and measures
	\[ X = [0,1], \, \mu_X = \Leb_{[0,1]} , \, Y = \real, \, \mu_Y \sim N(0,1) , \mu = \mu_X \times \mu_Y . \]
	Then Algorithm~\ref{algo:1} applied to data points distributed according to the push-forward of $\mu$ under $f$ should yield an approximation of $\bar{f} = \Ex^{\mu} f$, which according to \eqref{eqn:elctrc2} coincides with 
	$\bar{f} = \lambda$. Also note that $\rho(x)$, which is the standard deviation of the conditional measure $\mu(:|x)$, is itself a conditional expectation, namely, $\Ex^{\mu} \abs{ f - \bar{f} }^2 = \rho$. See Figure~\ref{fig:principal} for the results of our algorithm applied to data.

	\subsection{Conclusions} \label{sec:conclus}
	
	These two relatively simple problems from the fields of image denoising and manifold learning were a test for Algorithm~\ref{algo:1}, and the convergence guarantees of Theorems \ref{thm:1} and \ref{thm:2}. This also leads to several future directions of work, for a finer performance of numerical methods. 
	\begin{enumerate}
		\item Tuning parameter $\delta$ : The main challenge in the scheme based on \eqref{eqn:scheme} is to average out the effect of the $Y$-variable, within the integrals
		\[ \paran{Q^\mu f}(z) = \int_{X\times Y} p(z, x) f(x, y) d\mu(x,y) \approx \int_{B(z')} p(z, x) \int_Y f(x, y) d\mu(y|x) d\mu_X(x) . \]
		Here is $\delta'$ is the effective radius of integration, which goes to zero as $\delta$ goes to zero. The smaller $\delta$ is, the more number of total samples are needed so that sufficiently many of them fall with this sphere $B(z')$. Thus a larger smoothing radius $\delta$ leads to a faster convergence with $N$. On the other hand, a larger $\delta$ and $N$ also increases the condition number of the matrix $\Matrix{P}$, which could adversely affect the accuracy of the solution to the linear least squares problem \eqref{eqn:scheme3}. There is no recipe for tuning these parameters that would work uniformly well in all applications. 
		\item Images with limited resolution : Given a fixed resolution for an image, say $m\times n$ pixels, one would be limited to $N=mn$ data samples, as each pixel would correspond to a data point $x_n$ drawn uniformly from the unit square. As seen in Figure~\ref{fig:as3d}, this could lead to a deterioration in performance as the image function becomes more oscillatory. A remedy is to first increase the resolution of the image using the capability of RKHS for accurate out-of-sample evaluations. This would be the subject of a more thorough and focused study in a subsequent work.
		\item Subsampling : While it is desirable to use all available data samples when approximating $Q$, one has the flexibility of choosing a subset of the data-points for approximating $K^\mu$, as this is an integral operator on the lower dimensional space $X$. As long as the condition that the subsampling measure $\nu$ converges weakly to $\mu$, any subsampling strategy would suffice.
		\item Choice of kernel : The question of which kernel would be optimal in a learning problem has been a long standing question, with no clear and unique answer \citep[e.g.][]{narayan2021optimal, baraniuk1993signal, crammer2002kernel}. While we have explained the reason behind our choice of using diffusion kernels, various other adaptive kernels could be good candidates, such as variable bandwidth kernels \citep[e.g.][]{fan1995data, BerryHarlim2016} and dynamics adapted kernels \citep[e.g.][]{Giannakis2015, DasGiannakis_delay_2019}. Algorithm~\ref{algo:1} does not specify the kernel, and any RKHS-like kernel such as the diffusion kernel would be sufficient. 
		\item Finally, our results depend on the function $f$ having a continuous conditional expectation $\bar{f}$. This condition is violated images with background and foreground objects, as well as in audio streams involving human speech or ambient sounds. Adapting these situations to fit Assumptions~\ref{A:1}--\ref{A:4} is an interesting and promising direction of research.
	\end{enumerate}
	
	\section{Appendix}	
	\subsection{Proof of Lemma~\ref{lem:djhe09}} \label{sec:proof:djhe09}
	
	Fix an $\epsilon>0$ and an $x\in \support(X)$. We prove the lemma by finding a neighborhood of $x$ in $\support(\mu_X)$ such that for every $x'$ drawn from this neighborhood, $ \abs{ \int_Y f(x', \cdot) d\mu_{x'} - \int_Y f(x, \cdot) d\mu\rvert_{x} }<2\epsilon$. By Assumption~\ref{A:2} there is a neighborhood $U$ of $x$ in $X$ such that 
	\[ \norm{ f(x,\cdot) - f(x',\cdot) }_{C(Y)} < \epsilon, \quad \forall x'\in U . \]
	Next since $f\in C(X\times Y)$ there is a neighborhood $V$ of $x$ in $X$ such that
	\[ \abs{ \int f(x, \cdot) dm(x) - \int f(x, \cdot) dm(x')  } < \epsilon , \quad \forall x' \in V\cap \support(\mu_X) . \]
	Fix an $x'\in U\cap V \cap \support(\mu_X)$. Then
	\[\begin{split}
		& \abs{ \int_Y f(x', \cdot) d\mu_{x'} - \int_Y f(x, \cdot) d\mu\rvert_{x} } = \abs{ \int_Y f(x', \cdot) dm(x') - \int_Y f(x, \cdot) dm(x) } \\
		& \quad \quad = \abs{ \int_Y \SqBrack{ f(x', \cdot) - f(x, \cdot) } dm(x') - \int_Y f(x, \cdot) d \SqBrack{ m(x') - m(x) } }\\
		& \quad \quad \leq \int_Y \abs{ f(x', \cdot) - f(x, \cdot) } dm(x') - \abs{ \int_Y f(x, \cdot) d \SqBrack{ m(x') - m(x) } } \\
		& \quad \quad < 2\epsilon .
	\end{split}\]
	This completes the proof of Lemma~\ref{lem:djhe09}. \qed
	
	\subsection{Proof of Lemma~\ref{lem:ue93}} \label{sec:proof:ue93}
	
	Equations \eqref{eqn:def:A_oprtr} and \eqref{eqn:A4} together give :
	\[\begin{tikzcd}
		& L^2(\alpha_X) & C(X) \arrow[l, "\iota_\alpha"'] \\
		L^2(\nu) \arrow{dr}[swap]{ \tilde{K}^\nu } \arrow[r, "K^\nu"'] \arrow{ru}{ A_{\alpha, \nu} } & C(X) \arrow[r, "\iota_\alpha"'] \arrow{d}{ \iota_\nu } & L^2(\alpha_X) \arrow[bend right=10]{lu}{ \tilde{P}^{\alpha_X} } \arrow[u, "P^{\alpha_X}"'] \\
		& L^2(\nu) \arrow{ur}[swap]{ j_{\nu\to\alpha} }
	\end{tikzcd}\]
	Since the kernel is s.p.d., $\tilde{K}^\nu$ is an invertible operator. Since $\nu << \alpha$, the map $j_{\nu\to\alpha}$  is injective. By its Markovian property, $P^{\alpha_X}$ is also injective. Finally, the restriction $\iota_\alpha : C(X) \to L^2(\alpha_X)$ is also injective Thus their composition $\iota_\alpha P^{\alpha_X} j_{\nu\to\alpha} \tilde{k}^\nu$ is also surjective. This equals $A_{\alpha, \nu}$, which also must be surjective. 
	
	According to \eqref{eqn:def:A_oprtr}, $A_{\alpha, \nu}$ is the composition of the integral operator $\tilde{P}^{\alpha_X}$ along with the bounded operators $\iota_\alpha$ and $K^\nu$. This makes $A_{\alpha, \nu}$ a compact operator. \qed
	
	\subsection{Proof of Lemma~\ref{lem:inr0}} \label{sec:proof:inr0}
	
	Since by Lemma~\ref{lem:ue93} $A_{\alpha, \nu}$ is compact, by Shauder's theorem, its adjoint $A_{\alpha, \nu}^*$ is compact too. By   \eqref{eqn:def:B_oprtr}, $B_{\alpha, \nu, \epsilon}$ is the composition of $A_{\alpha, \nu}^*$ with the inverse of $A_{\alpha, \nu}^* A_{\alpha, \nu} + \epsilon$. Now $A_{\alpha, \nu}^* A_{\alpha, \nu} + \epsilon$ is a symmetric positive definite, bounded operator, whose spectrum lies in $[\epsilon, \infty)$. This makes its inverse bounded. Thus $B_{\alpha, \nu, \epsilon}$ is compact too.
	
	Next, by Lemma~\ref{lem:ue93}, $A_{\alpha, \nu}$ has the SVD
	\[ A_{\alpha, \nu} = \sum_{n=1,2,\ldots} \sigma_n \ket{u_n} \bra{v_n} , \]
	where $\braces{ u_n }_{n=1,2,\ldots}$ is an orthonormal basis for $L^2(\alpha_X)$, $\braces{ v_n }_{n=1,2,\ldots}$ is an orthonormal basis for $\paran{ \ker  A_{\alpha, \nu} }^\bot$, and $\sigma_1 \geq \sigma_2 \geq \ldots$ are the singular values of $A_{\alpha, \nu}$. By Lemma~\ref{lem:ue93}, the space $\paran{ \ker  A_{\alpha, \nu} }^\bot$ is trivial. Thus $\braces{ v_n }_{n=1,2,\ldots}$ is an orthonormal basis for the entire  $L^2(\nu)$. In that case
	\[ A_{\alpha, \nu}^* = \sum_{n=1,2,\ldots} \sigma_n \ket{v_n} \bra{u_n} , \quad A_{\alpha, \nu}^* A_{\alpha, \nu} = \sum_{n=1,2,\ldots} \sigma_n^2 \ket{v_n} \bra{v_n} . \]
	Continuing to utilize this expansion, we get compact diagonal operators
	\[ A_{\alpha, \nu}^* A_{\alpha, \nu} + \epsilon = \sum_{n=1,2,\ldots} \paran{ \sigma_n^2 + \epsilon } \ket{v_n} \bra{v_n} , \quad \SqBrack{ A_{\alpha, \nu}^* A_{\alpha, \nu} + \epsilon }^{-1} = \sum_{n=1,2,\ldots} \frac{1}{ \sigma_n^2 + \epsilon } \ket{v_n} \bra{v_n} . \]
	As a result,
	\[ B_{\alpha, \nu, \epsilon} = \SqBrack{ A_{\alpha, \nu}^* A_{\alpha, \nu} + \epsilon }^{-1} A_{\alpha, \nu}^* = \sum_{n=1,2,\ldots} \frac{\sigma_n}{ \sigma_n^2 + \epsilon } \ket{v_n} \bra{u_n} , \]
	and
	\begin{equation} \label{eqn:od93}
		B_{\alpha, \nu, \epsilon}   A_{\alpha, \nu} =  \sum_{n=1,2,\ldots} \frac{\sigma_n^2}{ \sigma_n^2 + \epsilon } \ket{v_n} \bra{v_n} , \quad 	B_{\alpha, \nu, \epsilon}   A_{\alpha, \nu} - \Id_{L^2(\nu)} = - \sum_{n=1,2,\ldots} \frac{\epsilon}{ \sigma_n^2 + \epsilon } \ket{v_n} \bra{v_n} .
	\end{equation}
	%
	This completes the proof of the lemma. 	\qed
	
	\subsection{Proof of Lemma~\ref{lem:do3}} \label{sec:proof:do3}
	
	Let $\braces{ \alpha_N }_{N=1}^{\infty}$ be a sequence of measures in $\MM$ converging weakly to $\mu$. It has to be shown that the continuous functions $g_N := \Ex^{
		\alpha_N, \delta} f $ converge uniformly to $g = \Ex^{\mu, \delta} f$. By Assumption \ref{A:1}, $\support(\mu_X) = X$ and is a compact metric space. At this point we recall the following elementary result from Real Analysis :
	
	\begin{lemma} \label{lem:psd93k}
		Let $X$ be a compact metric space, and $g_n$ be a sequence of Lipschitz functions converging to another Lipschitz function $g$ pointwise on a dense subset of $X$. Then the $g_n$ converge uniformly to $g$.
	\end{lemma}
	
	In our case, our functions are uniformly Lipschitz by the following lemma
	
	\begin{lemma} 
		Let $\kappa$ be a kernel lying in the space $\Lip(X; C(X))$. Fix a Borel measure $\beta$ on $X$, and denote the corresponding integral operator by $\mathcal{K}^{\beta}$. Then any $\phi \in L^2(\beta)$, 
		\[ \abs{ \paran{\mathcal{K}^{\beta}_{\delta} \phi}(x) - \paran{\mathcal{K}^{\beta}_{\delta} \phi}(x') } 
		\leq \norm{\kappa}_{Lip} \dist(x,x') \norm{ \phi }_{L^2(\beta)} , \]
	\end{lemma}
	
	This completes the proof of Lemma~\ref{lem:do3}. \qed
	
	\subsection{Proof of Lemma~\ref{lem:kpod9} } \label{sec:prof::kpod9}
	
	Denote $h$ denote the function $\paran{ \tilde{K}^\nu }^{-1} \iota_\nu \Ex^{\mu, \delta} f$. Let $\braces{ \alpha_N }_{N=1}^{\infty}$ be a sequence of probability measures on $X\times Y$, converging weakly to $\mu$. The claim of the lemma can be restated as 
	\begin{equation} \label{eqn:dgdp09}
		\forall \theta>0, \; \exists N_0\in \num, \, \epsilon_0>0 \mbox{ s.t. } \; \forall N>N_0, \, \forall \epsilon \in (0, \epsilon_0), \; \norm{ \paran{ B_{\alpha_N, \nu, \epsilon} A_{\alpha_N, \nu} - \Id } h }_{L^2(\nu)} < \theta .
	\end{equation}
	In the proof of Lemma~\ref{lem:inr0}, the variables $\sigma_n, v_n$ that appear in \eqref{eqn:od93} depend on the measure $\alpha$. To indicate this dependency, we change the notation to $\sigma_{n, N}, v_{n, N}$. By the spectral convergence of $P^{\alpha_{X,N}}$ to $P^{\mu_X}$ ( \citep[e.g.][Prop 25]{DasGiannakis_delay_2019} \citep[e.g.][Prop 13]{VonLuxburgEtAl2008} ), the singular vectors and singular values of the operator $A_{\alpha, \nu}$  converges to $A_{\mu, \nu}$. Thus the $v_{N,n}$ are right singular vectors of $A_{\alpha_N, \nu}$. Similarly, we have right singular vectors $V_n$ of $A_{\mu, \nu}$. The function $h$ can be expanded expanded along both these bases as 
	\[ h = \sum_{n=1,2,\ldots} a_{n,N} v_{n,N}, \quad h = \sum_{n=1,2,\ldots} a_n v_{n}. \]
	Since $\Ex^{\mu, \delta} f$ is assumed to be in the RKHS, by \citep[][Thm 2.1]{DasGiannakis_RKHS_2018}, for each index $n$, $\lim_{N\to\infty} a_{n,N} = a_n$. At this point, we take note of the fact that $B_{\alpha_N, \nu, \epsilon} A_{\alpha_N, \nu}$ is bounded in norm by some constant $\Gamma$, for every $N\in\num$. Since $h$ has a bounded $L^2(\nu)$ norm, there are $M_0, N_0\in\num$ such that 
	\[ \sum_{n\geq M_0} \abs{ a_{n,N} }^2 < \frac{1}{2(\Gamma+1)}\theta, \quad \forall N>N_0 . \]
	Then by \eqref{eqn:od93} we have
	\[\begin{split}
		\paran{ B_{\alpha_N, \nu, \epsilon} A_{\alpha_N, \nu} - \Id } h = \epsilon \sum_{n<M_0} \frac{a_n}{\sigma_{n}^2 + \epsilon} v_{n,N} + \paran{ B_{\alpha_N, \nu, \epsilon} A_{\alpha_N, \nu} - \Id } \sum_{n\geq M_0} a_{n,N} v_{n,N} .
	\end{split}\]
	Note that the first sum on the RHS converges to zero. Thus there is an $\epsilon_0$ such that the first sum is less than $\theta/2$ in norm, for all $\epsilon\in (0, \epsilon_0)$. The norm of the second term is less than $\gamma/2$ by design. Thus the condition \eqref{eqn:dgdp09} holds, and completes the proof of the lemma. \qed
	
	\subsection{Proof of Theorem~\ref{thm:2}} \label{sec:proof:2}
	
	The timeseries $\paran{ x_n, y_n }_{n=1}^N$ leads to the following \emph{sampling} measures :
	\[ \mu_M := \frac{1}{M} \sum_{m=1}^{N} \delta_{x_m}, \quad \mu_N := \frac{1}{N} \sum_{n=1}^{N} \delta_{x_n} , \quad \bar{\mu}_N := \frac{1}{N} \sum_{n=1}^{N} \delta_{(x_n, y_n)} .\]
	These three measures respectively play the role of $\nu$, $\alpha_X$ and $\alpha$ from Assumption \ref{A:4}. Since $\nu$ is built from a subsample of the support for $\alpha_X$, Assumption \ref{A:4} is fulfilled. Thus all the criterion for  Theorem~\ref{thm:1}~(ii) are fulfilled, and \eqref{eqn:thm1b} applies. The equidistribution assumed in Assumption~\ref{A:5} implies that $\mu, M\mu_N$, $\bar{\mu}_N$ converges weakly to $\mu_X$, $\mu_X$ and $\mu$ respectively. With these choices of $\nu, \alpha$, $L^2(\nu)$ and $L^2(\alpha_X)$ are isomorphic to $\cmplx^N$. The integral operators $P$ and $K^\nu$ take the form of the $N\times N$ matrices $\Matrix{P}$ and $\Matrix{K}$. Equation \eqref{eqn:thm1b} thus takes the form of \eqref{eqn:thm2}. \qed
	
	\bibliographystyle{unsrt}
	\bibliography{\Path References,ref}
\end{document}